\documentclass[10pt,twocolumn,letterpaper]{article}

\usepackage[pagenumbers]{wacv} 

\usepackage{times}
\usepackage{epsfig}
\usepackage{graphicx}
\usepackage{cuted} 
\usepackage{amsmath}
\usepackage{amssymb}
\usepackage{booktabs}
\usepackage{overpic}
\usepackage{tabularx}
\usepackage{pifont}    
\usepackage{colortbl}
\definecolor{darkgreen}{rgb}{0.0, 0.5, 0.0}

\definecolor{wacvblue}{rgb}{0.21,0.49,0.74}
\usepackage[pagebackref,breaklinks,colorlinks,allcolors=wacvblue]{hyperref}

\def\wacvPaperID{109} 
\def\confName{WACV}
\def\confYear{2026}

\title{Towards Egocentric 3D Hand Pose Estimation in Unseen Domains
}

\author{
Wiktor Mucha$^{1,2}$ \quad Michael Wray$^{3}$ \quad Martin Kampel$^{1}$\\[0.5em]
$^{1}$Computer Vision Lab, TU Wien \quad $^{2}$SoftServe Inc., Kraków, Poland, \quad $^{3}$University of Bristol\\
[0.3em]
{\tt\small \{wiktor.mucha, martin.kampel\}@tuwien.ac.at, michael.wray@bristol.ac.uk}
}

\begin{document}
\maketitle

\begin{strip}
\vspace{-1.5cm}
    \centering
    \includegraphics[width=\textwidth]{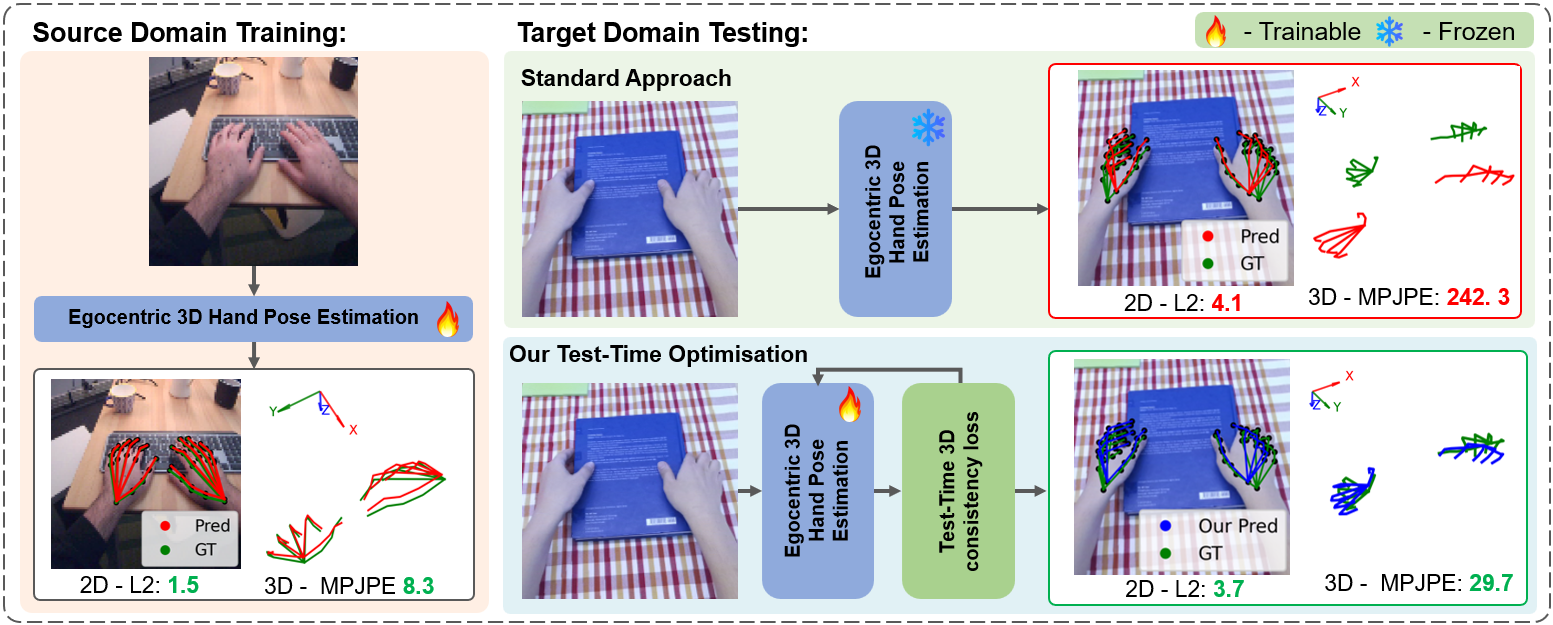} 
    \captionof{figure}
{\textbf{V-HPOT improves egocentric 3D hand pose estimation in cross-domain scenarios,} where a model is trained in source and tested in different target domain. We propose training the model to estimate the $z$-coordinate in virtual camera space to mitigate overfitting to source domain camera intrinsics. At test time, we optimise predictions using a novel self-supervised loss that enforces 3D spatial consistency, allowing the model to refine depth perception by aligning depth-augmented poses in 3D camera space. This significantly improves absolute 3D pose error (MPJPE) without any labelled data, outperforming existing methods focused solely on 2D improvement.}

    \label{fig:teaser}
\end{strip}

\begin{abstract}

We present V-HPOT, a novel approach for improving the cross-domain performance of 3D hand pose estimation from egocentric images across diverse, unseen domains. State-of-the-art methods demonstrate strong performance when trained and tested within the same domain. However, they struggle to generalise to new environments due to limited training data and depth perception -- overfitting to specific camera intrinsics. Our method addresses this by estimating keypoint z-coordinates in a virtual camera space, normalised by focal length and image size, enabling camera-agnostic depth prediction. We further leverage this invariance to camera intrinsics to propose a self-supervised test-time optimisation strategy that refines the model's depth perception during inference. This is achieved by applying a 3D consistency loss between predicted and in-space scale-transformed hand poses, allowing the model to adapt to target domain characteristics without requiring ground truth annotations. V-HPOT significantly improves 3D hand pose estimation performance in cross-domain scenarios, achieving a 71\% reduction in mean pose error on the H2O dataset and a 41\% reduction on the AssemblyHands dataset. Compared to state-of-the-art methods, V-HPOT outperforms all single-stage approaches across all datasets and competes closely with two-stage methods, despite needing $\approx\times3.5$ to $\times14$ less data. \url{https://github.com/wiktormucha/vhpot}.

\end{abstract}

\vspace{-0.5cm}
\section{Introduction}
\label{section:introduction}

Egocentric vision has emerged as a rapidly growing research area in computer vision, fueled by datasets including \textit{Epic-Kitchens}~\cite{Damen2018EPICKITCHENS} \textit{Ego4D}~\cite{grauman2022ego4d} or \textit{HOT3D}~\cite{banerjee2024hot3d}, with applications spanning assistive technologies ~\cite{text2taste} and robotics~\cite{fang2024egopat3dv2}. Since hands play a central role in the egocentric viewpoint, one of the key challenges is to accurately estimate the 3D hand pose of the camera wearer. A correct 3D hand pose representation enables interaction with virtual environments in VR/AR applications, facilitates action recognition, or can be used to transfer manipulation skills from human experts to robots in real-world scenarios~\cite{banerjee2024hot3d}.

Recent works tackle egocentric 3D hand pose estimation through two distinct approaches. Two-stage methods~\cite{liu2021semi, ohkawa:cvpr23, wildhands2025} first detect hand regions then estimate keypoints, allowing detector pre-training on large datasets without 3D labels. However, this introduces critical limitations: increased computational overhead, cascading failures when detection errors occur, and loss of crucial contextual information in hand-object interactions due to tight cropping~\cite{cho2023transformer}.
Single-stage methods~\cite{cho2023transformer, wen2023hierarchical, mucha2025sharp} overcome these limitations by directly estimating keypoints from full RGB frames in one forward pass, preserving scene context and enabling joint reasoning about hands and objects~\cite{hasson2019learning, karunratanakul2020grasping, ye2022s}, which is essential for accurately capturing the complex spatial relationships in interaction scenarios.

Despite these advancements, most methods only report performance within the same domain~\cite{Kwon_2021_ICCV, ohkawa:cvpr23, cho2023transformer, wen2023hierarchical, mucha2025sharp}, leaving performance in unseen domains unaddressed -- critical for real-world applications that must function robustly in uncontrolled environments. Models overfit to specific camera intrinsics and environments, causing performance degradation in novel settings. One possible solution is to train models with larger and more diverse datasets; however, acquiring 3D pose annotations is costly, requiring multi-camera setups~\cite{Kwon_2021_ICCV, ohkawa:cvpr23} or motion capture systems~\cite{banerjee2024hot3d} that constrain data collection to laboratories, limiting generalisation capabilities towards real-world data.

We address these challenges with V-HPOT (Virtual Camera-based Hand Pose Optimisation at Test-time) presented in Figure~\ref{fig:teaser}, a single-stage approach with two key components. First, we formulate z-coordinate prediction in Virtual Camera (VC)~\cite{brazil2023omni3d} space rather than metric space, making depth perception less dependent on camera intrinsics and enabling depth augmentations that enhance generalisation across datasets with varying focal lengths. During training, we incorporate an auxiliary pseudo-depth prediction task supervised by \textit{DPT-Hybrid}~\cite{ranftl2021vision}, improving cross-domain generalisation. Second, we introduce a novel self-supervised Test-Time Optimisation (TTO) strategy that dynamically adapts the model to previously unseen environments by refining the feature extractor through a consistency loss between predicted and depth-transformed hand poses in 3D space. This is only possible with our camera-agnostic depth representation, which maintains consistency under depth transformations, unlike conventional metric representations that suffer from scale-depth ambiguity.

V-HPOT achieves consistent improvements in cross-domain scenarios, reducing mean pose error by 71\% and 41\% for \textit{H2O}~\cite{Kwon_2021_ICCV} and \textit{AssemblyHands}~\cite{ohkawa:cvpr23} datasets, respectively. Compared to the state of the art, we outperform other single-stage methods in cross-domain settings on \textit{H2O}~\cite{Kwon_2021_ICCV} and \textit{AssemblyHands}~\cite{ohkawa:cvpr23} across all evaluation metrics. Additionally, we assess the 2D projections of our method on the in-the-wild \textit{Epic-Kpts} dataset~\cite{wildhands2025}, which lacks 3D annotations, where it surpasses all single-stage approaches and performs comparably with two-stage methods, despite our model being trained from $\approx \times 3.5$ to $\times14$ less data. Our key contributions are as follows:

\begin{itemize}

\item We adapt virtual camera space to hand pose estimation, reducing depth perception's dependency on camera intrinsics and enabling effective depth augmentations.

\item We propose pseudo-depth prediction as an auxiliary task for cross-domain 3D hand pose estimation.

\item We propose a test-time optimisation framework with a novel self-supervised 3D consistency loss for 3D hand pose estimation, leveraging an intrinsic invariant virtual camera space. It allows the model to refine depth perception in unseen domains without any labelled data by enforcing alignment between 3D depth-augmented poses.

\item{
We demonstrate V-HPOT effectiveness through extensive experiments across multiple dataset domains, achieving substantial cross-domain performance improvements: 71\% error reduction on the RGB \textit{H2O Dataset} and 41\% on the monochromatic \textit{AssemblyHands}.
}

\end{itemize}

\section{Related Work}

\label{sec:related_work}


\paragraph{Egocentric 3D Hand Pose Estimation.}

Hand pose estimation in egocentric vision is challenging due to self-occlusion, limited field of view, and varying perspectives. Current approaches can be categorised into two main paradigms:
Two-stage methods ~\cite{rong2021frankmocap,park2022handoccnet,wildhands2025,hamerpavlakos2024reconstructing} first localise hand regions before estimating poses. FrankMocap~\cite{rong2021frankmocap} employs a convolutional backbone and an HMR-style~\cite{kanazawa2018learning} decoder to predict MANO parameters under 2D and 3D supervision. HandOccNet~\cite{park2022handoccnet} integrates an FPN~\cite{lin2017feature} backbone and transformer modules for heatmap intermediate representations. HaMeR~\cite{hamerpavlakos2024reconstructing} leverages a ViT~\cite{dosovitskiy2020image} with a transformer decoder. WildHands~\cite{wildhands2025} incorporates image features and intrinsics-aware positional encodings for each hand crop, before regressing 3D hand pose. Auxiliary supervision comes from hand segmentation masks and grasp labels in the \textit{Epic-Kitchens} dataset. While effective, these approaches introduce computational overhead and lose crucial interaction context due to cropping~\cite{cho2023transformer} and train on a large amount of data.

In contrast, single-stage methods estimate hand pose directly from entire images~\cite{tekin2019h+,Kwon_2021_ICCV,mucha2025sharp,fan2023arctic,cho2023transformer,wen2023hierarchical}. Tekin et al.~\cite{tekin2019h+} predict a 3D pose grid from a single RGB image, while Kwon et al.~\cite{Kwon_2021_ICCV} extend this to both hands. Transformer-based models have also been explored, with Cho et al.~\cite{cho2023transformer} focusing on frame-wise 3D reconstruction and Wen et al.~\cite{wen2023hierarchical} addressing occlusions with a sequence-based approach. ArcticNet-SF~\cite{fan2023arctic} combines a convolutional backbone with an HMR-style decoder for hand and object pose estimation. SHARP~\cite{mucha2025sharp} introduces pseudo-depth estimation to isolate hands and objects in egocentric scenes, leveraging arm distance for improved accuracy. Many of these methods extend to 3D object pose estimation, where joint reasoning about hands and objects is proven to enhance reconstruction of both~\cite{hasson2019learning, karunratanakul2020grasping, ye2022s}.

Despite advancements, most studies focus on controlled settings~\cite{ohkawa:cvpr23,Kwon_2021_ICCV,cho2023transformer,wen2023hierarchical,tekin2019h+}, with limited attention to real-world egocentric data~\cite{hamerpavlakos2024reconstructing, wildhands2025}. Building on single-stage approaches~\cite{Kwon_2021_ICCV,cho2023transformer,wen2023hierarchical}, we evaluate 3D hand pose in a cross-domain setting with different sensors following~\cite{wildhands2025}. Compared to~\cite{wildhands2025}, our method improves the performance of single-stage approaches, relies on a single training dataset, minimises the number of auxiliary tasks to one introducing a novel one, and directly improves depth estimation ($z$-coordinate) from a single RGB image. 

\vspace{-0.5cm}
\paragraph{Test-time optimisation for pose estimation.}

Test-time optimisation improves model performance during inference by adapting to test data in an unsupervised manner~\cite{sun2020test, hu2024fast, peng2023source}. This adaptation occurs online, either by optimising an auxiliary task~\cite{sun2020test} or passively through non-training methods such as statistical adaptation of Batch Normalisation layers~\cite{li2016revisiting}. 
Test-time adaptation for pose estimation is achieved via self-training and self-supervision tasks. Li et al.~\cite{li2021test} use multi-view images of the same person to synthesise the input image based on initial pose prediction. Hu et al.~\cite{hu2024fast} employ initial pose predictions to mask image regions and apply inpainting as an auxiliary task. Hasson et al.~\cite{hasson2020leveraging} perform weakly supervised learning with differentiable rendering and optical flow for temporal consistency in 3D hand-object reconstruction, while~\cite{cao2021reconstructing} refine hand-object alignment using additional knowledge regarding 3D contact priors and collision avoidance.

Our work differs from existing methods as we focus on 3D hand pose estimation, rather than simply refining 2D predictions~\cite{li2021test, hu2024fast, peng2023source}. Current 3D hand pose adaptation approaches either rely on additional ground truth data~\cite{hasson2020leveraging} or require object interaction cues~\cite{cao2021reconstructing}, which limits their applicability in non-interaction scenarios. In contrast, we propose a novel self-supervision for enhancing depth perception, leading to more robust and generalisable 3D hand pose estimation across domains, independent of camera intrinsic, without the need for additional ground truth labels.

\vspace{-0.5cm}
\paragraph{Self-supervision in hand pose.}

TTO heavily depends on selecting an unsupervised task aligned with the main objective. Various self-supervised approaches exist for hand pose estimation. Some generate pseudo-labels using 2D pose estimators \cite{zheng2023hamuco, chen2021model} or utilise 2D ground truth poses \cite{chen2021temporal}, refining them to improve 3D representations through cross-sample consistency. HaMuCo~\cite{zheng2023hamuco} leverages multi-view consistency during training to enhance single-view estimates. TASSN \cite{chen2021temporal} enforces temporal consistency between forward and reverse video frames, ensuring stable 3D predictions from single-view ground truth 2D labels. For single-view, non-temporal approaches, S2-Hand \cite{chen2021model} refines noisy 2D labels through rendering-based self-supervision, while Spurr et al. \cite{spurr2021self} apply contrastive learning with geometric and appearance-based augmentations between similar frames, enforcing consistency in the 2D space, while Yang et al.~\cite{yang2021semihand} utilise rotations. Lin et al. use contrastive pretraining by mining similar hand representations~\cite{lin2025simhand}.

Our work introduces a novel self-supervised task specifically for TTO, unlike prior approaches focused on pre-training. We neither rely on pseudo-labels~\cite{zheng2023hamuco, chen2021model,lin2025simhand}, temporal cues~\cite{spurr2021self}, nor rendering~\cite{chen2021model}. While we leverage consistency between similar poses~\cite{spurr2021self, chen2021temporal}, we avoid multi-frame dependencies~\cite{chen2021temporal} or data mining~\cite{lin2025simhand} not suitable for test-time. Additionally, we extend beyond image-based augmentations~\cite{spurr2021self} by applying transformations directly in 3D space to augment hand poses and force 3D consistency.

\section{Method}
\label{sec:methodology}

\begin{figure}[t]
  \centering
  \includegraphics[width=\linewidth]{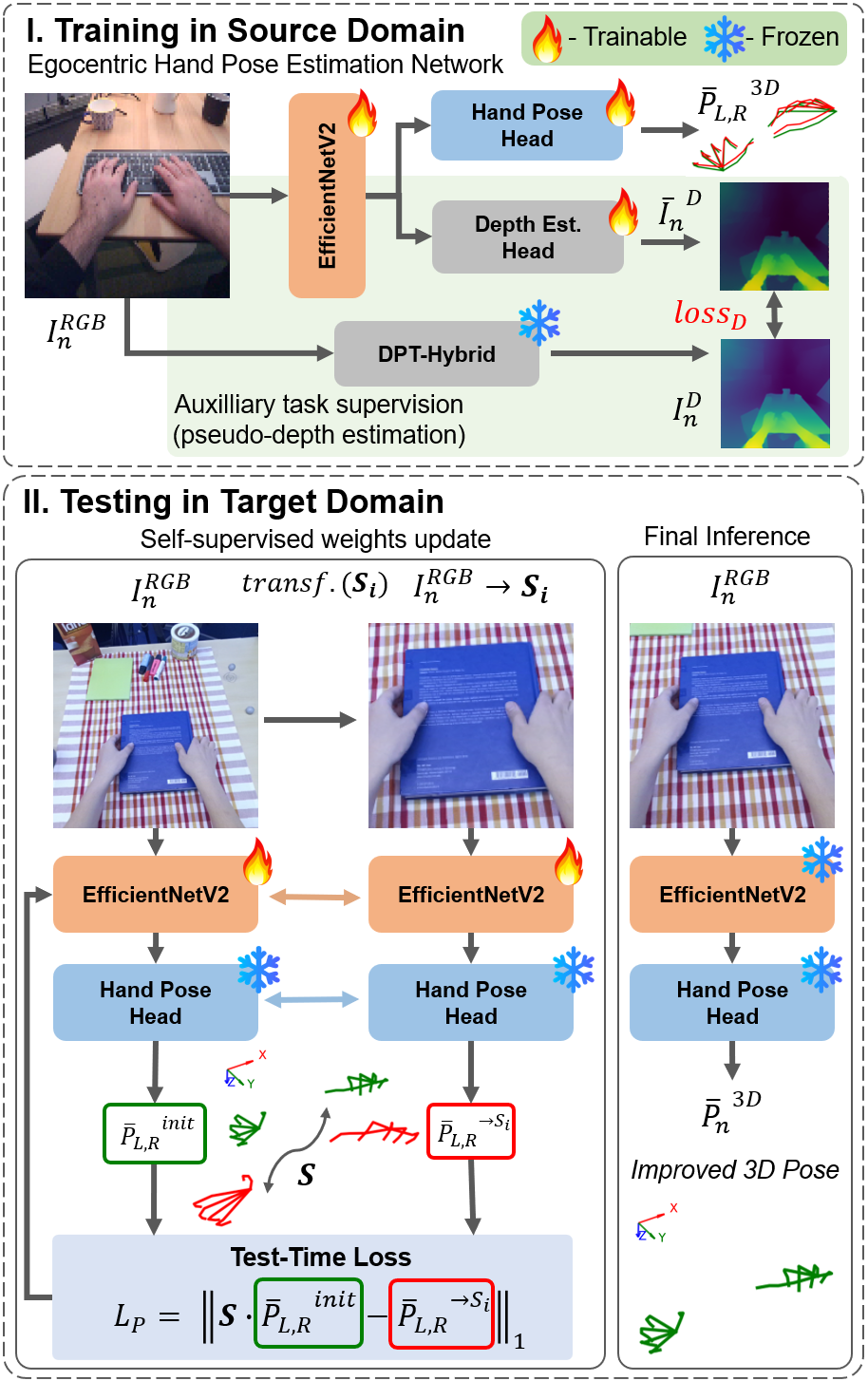}
  \caption[Caption for LOF]{\textbf{V-HPOT:} We train in source domain to estimate 3D hand pose \( P_{L,R}^{3D} \) from an image \( I_n^{RGB} \) in virtual camera space, with an auxiliary pseudo-depth task \( I_n^D \) supervised by \textit{DPT-Hybrid}~\cite{ranftl2021vision}. At test time, the model predicts an initial pose \( P_{L,R}^{\text{init}} \) and applies depth augmentation by scaling \( I_n^{RGB} \) with \( S_i \) in virtual camera space, yielding \( P_{L,R}^{\rightarrow S} \). A 3D consistency loss between these poses refines the backbone, enhancing final predictions.

  }
  \label{fig:method_train}
  \vspace{-0.5cm}
\end{figure}

Our method introduces two key components to improve the egocentric 3D hand pose estimation (see Figure~\ref{fig:method_train}). (1) We train the model to estimate depth ($z$-coordinate) in virtual camera space, making predictions agnostic to camera intrinsic variations across datasets while introducing pseudo-depth estimation as an auxiliary task. (2) We propose a test-time training pipeline that enhances the model's adaptability during inference by incorporating a 3D consistency loss that leverages various distance augmentations in the virtual camera space, enabling the model to learn robust hand pose estimation in new domains in a fully unsupervised manner.

\subsection{3D Hand Pose Estimation Network}

Our egocentric hand pose estimation network adopts a single-stage architecture. Following standard practice, we represent each hand with 21 keypoints $j\!\in\!J$ denoting joint positions, expressed as $\mathit{P}^{3D}_{L,R}(x,y,z), \in \mathbb{R}^{J \times 3}$ where $L,R$ stands for the left or right hand and $x, y, z$ represents the point $j$ in 3D. We follow a common practice, where $x, y$ are estimated via heatmaps and $z$ is regressed~\cite{mucha2025sharp}. 

Given an input RGB frame \( I_n^{RGB} \in \mathbb{R}^{3\times 224\times 224} \), we extract features using \textit{EfficientNetV2-S}~\cite{tan2021efficientnetv2} to produce a feature map $F_M$. This backbone is selected following experiments in~\cite{mucha2024my}, where it outperforms other architectures in the egocentric hand pose task. $F_M$ is then fed into two independent upsamplers—one per hand—to generate 2D keypoint heatmaps $H_{L,R}$. Per-hand upsamplers are first presented in InterHands~\cite{moon2020interhand2}, a pioneering study focused on interacting hands, and became common in egocentric hand pose~\cite{Kwon_2021_ICCV,cho2023transformer,wen2023hierarchical,mucha2025sharp} due to improvements in pose estimation during hand interactions. Each upsampler consists of four transposed convolution layers with batch normalisation and ReLU , followed by a pointwise convolution in the final layer. The output, \( \mathit H_{L,R} \in \mathbb{R}^{J\times 112\times 112} \), represents per-joint probability distributions, which are converted into 2D keypoints ($P^{2D}_{L,R}, \in \mathbb{R}^{J\times 2}$). Depth (\( z \)) is predicted separately via a Multi-Layer Perceptron (MLP) (\( MLP^{Z}_{L,R} \)), and combined with \( \mathit P^{2D}_{L,R} \) to form 2.5D keypoints (\( \mathit P^{2.5D}_{L,R} \)). To obtain the 3D hand pose in camera space, we apply the pinhole camera model (Equation \ref{eq:pinhole}), transforming \( \mathit P^{2.5D}_{L,R}(x,y,z) \) into \( \mathit P^{3D}_{L,R} \) using intrinsic parameters $K$. 

\begin{equation}
    \mathbf{P}^{3D}_{L,R, j} = z_j \cdot K^{-1} \begin{bmatrix} x_j \\ y_j \\ 1 \end{bmatrix}, \quad \forall j \in J
    \label{eq:pinhole}
\end{equation}

Additionally, hand presence is determined via handedness modules (\( MLP^H_{L,R} \)), predicting the probability of each hand being present (\( h_L, h_R \in \mathbb{R}^2 \)) in the image $I_n^{RGB}$.  

\subsection{Understanding depth in new domains}
\paragraph{Introducing virtual camera space.}

Estimating metric depth from pixel values requires an implicit mapping of 2D keypoints to 3D distances, which becomes ambiguous due to variations in camera intrinsics. Since camera intrinsics differ across datasets, models trained on a single dataset tend to overfit to specific depth representations, leading to poor generalisation in cross-domain scenarios where data is often captured by different sensors.

To address this issue, we implement a concept of a \textit{virtual camera (VC)}~\cite{brazil2023omni3d} for hand pose estimation. This approach transforms metric depth values $z$ to a uniform virtual camera space, denoted as $z_v$. The transformation is achieved by scaling depth values $z$ using known camera intrinsics, ensuring a consistent, effective image size and focal length across datasets. We denote these standardised parameters as $H_v$ and $f_v$, which are treated as hyperparameters. Following this, virtual depth is defined as the transformation of ground truth $z$-coordinate $z_v = z \cdot \frac{f_v}{f} \frac{H}{H_v}$ where $H$ and $f$ are source image height and focal length. The model learns to predict $\hat{z}_v$, which is scaled back to metric $z$ for the output. Additionally, leveraging virtual camera space enables scale augmentations that enhance both 2D and 3D pose estimation. Without it, augmentations would improve 2D representation but degrade 3D understanding~\cite{brazil2023omni3d}.

\paragraph{Training objectives.}

Our objectives are 3D hand pose estimation and pseudo-depth estimation as an auxiliary task. The prediction of a pseudo-depth map is supervised using the prediction by the \textit{DPT-Hybrid}~\cite{ranftl2021vision}. We use an L1 loss over all pixels $N$ to minimise the difference between the predicted depth $\hat{I}^D$ and the reference pseudo-depth $I^{D}$:
\begin{equation}
    \mathcal{L}_{\text{d}} = \frac{1}{N} \sum_{i=1}^{N} \left| \hat{I}^D_i - I^D_i \right| \quad 
\end{equation}
For 2D hand keypoint localisation, we supervise the network using an Intersection over Union (IoU) loss over the predicted heatmaps. Given the predicted heatmap $\hat{H}$ and the ground truth heatmap $H$, the IoU loss is defined as:
\begin{equation}
    \mathcal{L}_{xy} = 1 - \frac{\sum_{j=1}^{21} \min(\hat{H}_j, H_j)}{\sum_{j=1}^{21} \max(\hat{H}_j, H_j)}
\end{equation}
For 3D hand pose estimation, we directly optimise the obtained depth ($z_v$-coordinate) using an L1 loss:
\begin{equation}
    \mathcal{L}_{z_v} = \sum_{j=1}^{21} | \hat{z}_{v,j} - z_{v,j} |
\end{equation}
To predict whether the image contains left, right, or both hands, we use a handedness classification module. The handedness prediction $\hat{h}$ is supervised using the cross-entropy loss with the ground truth label $h$:
\begin{equation}
    \mathcal{L}_{h} = - \sum_{c \in \{\text{left}, \text{right}\}} h_c \log \hat{h}_c
\end{equation}
where $h$ is a one-hot encoded ground truth vector indicating whether the hand is left or right. The final objective is: 
\begin{equation}
    \mathcal{L}_{train} = 
    \lambda_{\text{d}} \mathcal{L}_{\text{d}} + 
    \lambda_{xy} \mathcal{L}_{xy} + 
    \lambda_{z_v} \mathcal{L}_{z_v} + 
    \lambda_{h} \mathcal{L}_{h}
\end{equation}
where $\lambda_{\text{d}}, \lambda_{\text{xy}}, \lambda_{z_v}, \lambda_{\text{h}}$ are hyperparameter weights controlling the relative importance of each term.

\paragraph{Generalising to new domains at test time.}

We propose a fully self-supervised test-time optimisation pipeline that adapts hand pose estimation to unseen environments without requiring target domain annotations. This adaptation during inference is only possible because of our virtual camera transformation, which creates a camera-intrinsic-uniform representation space.

Specifically, we introduce a consistency loss to improve 2D and depth generalisation. The model first infers the 3D hand pose in virtual camera space, producing an initial prediction \( \mathbf{P}_{L,R}^{\text{init}} \). We then apply \( n = 2 \) independent random scaling factors \( S_i \sim \mathcal{U}(1.0, 1.25) \) to augment the depth, generating \( n\) transformed poses \( \mathbf{P}_{L,R}^{\rightarrow S_i} \) that simulate shifts in camera distance. These transformed poses are subsequently passed through the network again, producing new predictions. Since the spatial relationships between joints should remain consistent despite the depth transformation, we enforce this 3D consistency via the L1 loss function:

\begin{equation}
\mathcal{L}_{\text{test-time}} = \sum_{i=1}^{n} \left\| S_i \cdot \mathbf{P}_{L,R}^{\text{init}} - \mathbf{P}_{L,R}^{\rightarrow S_i} \right\|_1
\end{equation}

This consistency loss optimises the feature extractor weights while keeping the hand pose head weights frozen. It is carried out \textit{online} using the first $F_N$ frames of the test data. This choice ensures a balance between quick adaptation and avoiding overfitting, allowing the model to adjust to domain-specific variations without compromising generalisation. Empirical observations show that using a larger fraction leads to diminishing returns, while a smaller fraction results in insufficient adaptation. Initially, the model is trained on the source dataset and undergoes a one-time weight initialisation. During testing, the weights are iteratively refined using \( \mathcal{L}_{\text{test-time}} \) before making predictions in the same manner for every domain. After processing the first $F_N$ of the data, optimisation is disabled, and the model proceeds with inference without any further updates.

\section{Experiments}

\label{results}

We conduct cross-domain evaluation: training on source dataset, testing on target dataset from different domain.   

\subsection{Experimental Methodology}

\paragraph{Datasets.} 

Our study employs four egocentric datasets: three from laboratory settings and one collected in the wild. For lab datasets, we use: (1) \textit{H2O}~\cite{Kwon_2021_ICCV}, which provides multi-view RGB-D recordings with 21 keypoint annotations per hand, 6D object poses, and scene metadata (55K training, 23K testing frames); (2) \textit{AssemblyHands}~\cite{ohkawa:cvpr23}, a large-scale monochrome dataset (380K training, 32K validation frames); and (3) \textit{HOT3D}~\cite{banerjee2024hot3d}, featuring multi-view RGB video streams of 19 subjects interacting with 33 objects (170K training, 42K testing frames). For in-the-wild evaluation, we use \textit{Epic-Kpts}~\cite{wildhands2025}, a subset of Epic-Kitchens~\cite{Damen2018EPICKITCHENS} with 2D annotations for 21 hand joints from 4K images sampled from VISOR split~\cite{VISOR2022}. For this dataset, we project our 3D predictions into 2D for assessment. Specific dataset combinations for experiments are detailed in later sections.

\vspace{-0.4cm}
\paragraph{Metrics.} We use standard hand pose estimation metrics: Mean Per Joint Position Error (MPJPE), measuring Euclidean distance (mm) between predicted and ground-truth 3D keypoints across 21 hand joints; root-aligned MPJPE (MPJPE-RA) (mm) with poses aligned by root joint subtraction; Mean Relative-Root Position Error (MRRPE) (mm), the quantifying distance between the left and right hand root joints and measuring absolute pose information; and L2 error (pixels) between ground truth 2D keypoints and 2D projections of predicted 3D keypoints on 224×224 input images. All 3D metrics are computed in camera space.

\vspace{-0.4cm}
\paragraph{Experimental Setup.}

Each run, including ablation studies, is repeated three times to mitigate the impact of random network initialisation. To enable comparison with related work~\cite{wildhands2025,cho2023transformer, hamerpavlakos2024reconstructing, rong2021frankmocap}, the model is trained on \textit{HOT3D}~\cite{banerjee2024hot3d} as source dataset and evaluated on \textit{H2O Dataset}~\cite{Kwon_2021_ICCV} and \textit{AssemblyHands}~\cite{ohkawa:cvpr23}. A cross-domain experiment showing improvements in all scenarios is available in the Supp.~\ref{supp:sec:cross-domain}. The optimisation is performed using Stochastic Gradient Descent (SGD) with a learning rate of \(l_r = 0.1 \) and momentum \( m = 0.9 \). The learning rate is progressively reduced by a factor of \( \alpha = 0.5 \) every five epochs, starting from the 25th epoch, where the process ends at 50 epochs. Training loss $\mathcal{L}_{train}$ parameters are set up following~\cite{mucha2025sharp}. For virtual camera parameters, we select constant $H_v = 720$ and $f_v=512$ following~\cite{brazil2023omni3d}. Data augmentation includes horizontal flipping, resizing, and appearance-based transformations such as blurring and de-pixelization. The batch size is set to \( b_s = 32 \). Model weights are saved based on the lowest MPJPE observed in the validation set.

During test, we optimise \( \mathcal{L}_{\text{test-time}} \) using SGD with \( l_r = 0.3 \) and \( m = 0.2 \). For experiments, we use $F_N$ equal 5\% of test data, whilst for comparison with other methods we use a fixed number (5\% of \textit{H2O} - 960 frames) to simulate realistic deployment scenarios. The model performs best when enforcing consistency loss across $n = 2$ shifted poses.

\subsection{Results}

We present ablations on V-HPOT's key components: VC, TTO, auxiliary task, and TTO loss. Finally, we compare V-HPOT with the state of the art in cross-domain scenarios.

\vspace{-0.4cm}
\paragraph{Virtual Camera and Test-Time Optimisation.}

We evaluated our VC approach and TTO impact through ablations:

\begin{enumerate}

\item \textit{base}: Training and testing without VC transformation
\item \textit{VC}: Applying VC transformation during training for camera intrinsic robustness, without TTO

\item \textit{TTO}: Incorporating TTO without VC

\item \textit{VC+TTO}: Integrating both VC transformation and TTO, enabling intrinsic-independent depth learning with adaptive inference (our complete V-HPOT architecture)

\end{enumerate}

\begin{figure}[t]
  \centering
  \includegraphics[width=\linewidth]{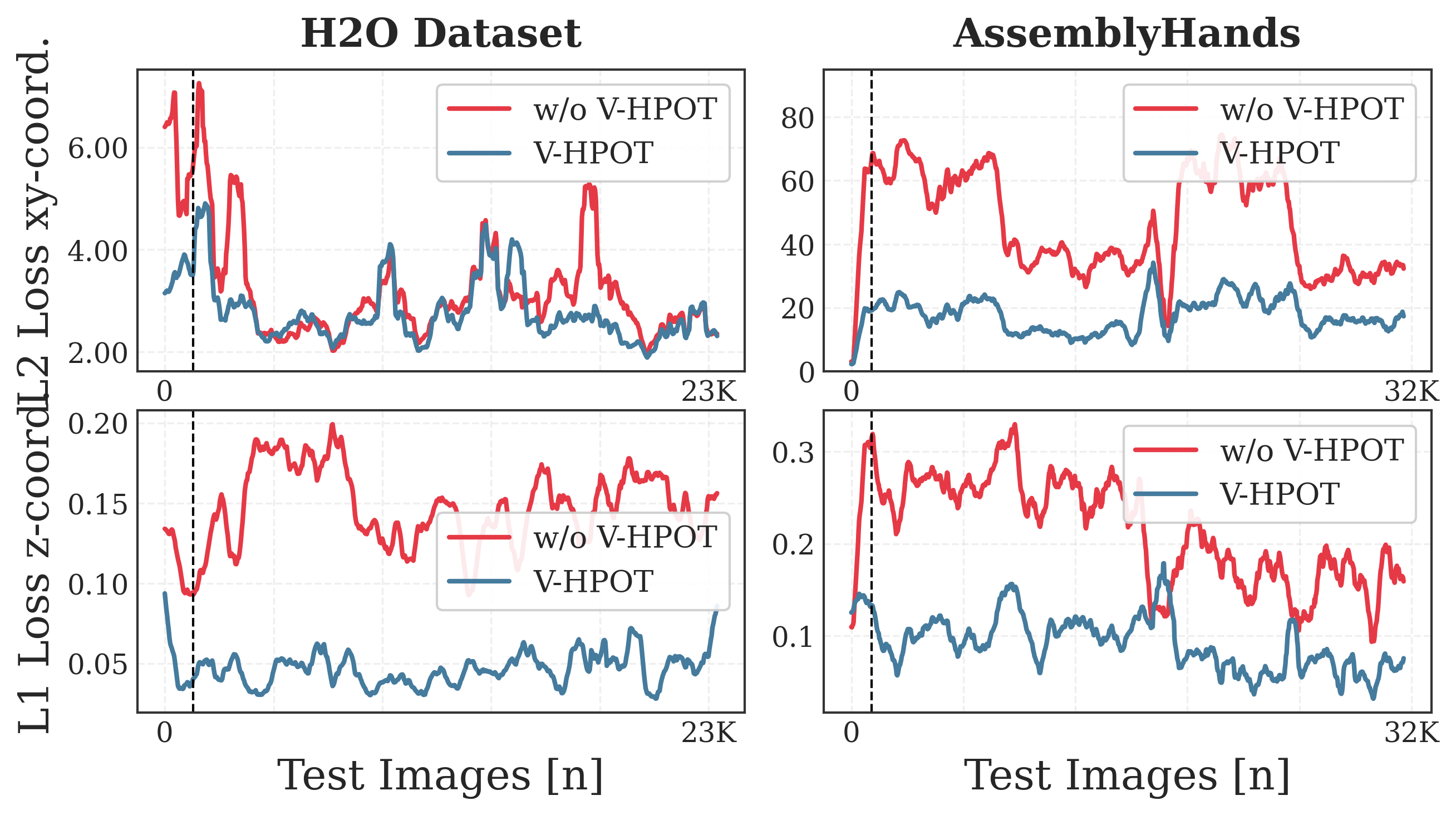}
\caption{\textbf{Visualisation of L2 loss for \(x, y\) and L1 loss for the \(z\) coordinate over test samples in cross-domain scenarios}. The dashed line represents 5\% of data where V-HPOT stops optimising. Continuous improvement is observed for most samples.}
  \label{fig:loss_diff}
  \vspace{-0.25cm}
\end{figure}

Results in Table \ref{tab:ablation_vc} demonstrate that V-HPOT \textit{(VC+TTO)} yields superior performance across all metrics, reducing MPJPE by 70\% on H2O and 41\% on AssemblyHands (AsHa). The combined VC and TTO enhances depth perception in unseen domains by addressing key limitations: \textit{TTO} alone leads to source camera intrinsic overfitting, while \textit{VC} alone lacks a feedback mechanism for weight adjustment. Both target datasets exhibit improvements with our best \textit{(VC+TTO)} in 3D error, but gains vary by dataset. \textit{AssemblyHands}~\cite{ohkawa:cvpr23} showed greater 2D L2 reduction, likely due to its monochromatic nature offering more improvement potential compared to the source RGB (HOT3D). In \textit{Epic-Kpts} L2 reduces by 30\% (12.9-8.4). Figure \ref{fig:loss_diff} visualises coordinate-specific losses (L2 for $x,y$; L1 for $z$) on target datasets, comparing \textit{base} model with \textit{(V-HPOT)}. Results show consistent improvements across images, even with high error in initial prediction.

\vspace{-0.4cm}
\paragraph{Auxiliary pseudo-depth estimation.} We evaluate its impact on training in Table \ref{tab:ablation_aux} by comparing the \textit{V-HPOT} against \textit{V-HPOT-D\textsubscript{tr}}, a version with excluded auxiliary pseudo-depth estimation task (D\textsubscript{tr}) during training and with \textit{V-HPOT(DA)} where we utilise DepthAnything~\cite{yang2024depth} instead of \textit{DPT-Hybrid}. Results show that including D\textsubscript{tr} improves cross-domain performance across all metrics and datasets. Performance between \textit{V-HPOT(DA)} and \textit{V-HPOT} varies, thus we select \textit{DPT-Hybrid} as it achieves superior absolute 3D error and faster inference (26.8ms vs 39.8ms mean of 1000 trials on RTX3090). Qualitative observations show that DepthAnything estimates finer background details, whilst hands remain similar (Supp.~\ref{supp:sec:qual}).

{
\setlength{\tabcolsep}{0.0pt}

\begin{table}[t]
    \centering
    \small
    \caption{\textbf{Ablation study presenting impact} of proposed virtual camera transformation and TTO strategy. We observe an improvement of V-HPOT \textit{(VC+TTO)} over baseline in every metric.}
    \label{tab:ablation_vc}
    \begin{tabularx}{\linewidth}{l>{\centering}X>{\centering}X>{\centering}X>{\centering}X>{\centering}X>{\centering}X>{\centering}X>{\centering\arraybackslash}X}
        \toprule
        & \multicolumn{2}{c}{{MPJPE} $\downarrow$} & \multicolumn{2}{c}{{MPJPE-RA} $\downarrow$} & \multicolumn{2}{c}{{MRRPE} $\downarrow$} & \multicolumn{2}{c}{{L2} $\downarrow$} \\
        \cmidrule(lr){2-3}
        \cmidrule(lr){4-5}
        \cmidrule(lr){6-7}
        \cmidrule(lr){8-9}
        & H2O & AsHa & H2O & AsHa & H2O & AsHa & H2O & AsHa \\
        \midrule
        \textit{base} & 179.6 & 297.7 & 52.7 & 105.3 & 126.2 & 301.8 & 7.4 & 80.1 \\
        \textit{VC} & 146.1 & 302.3 & 63.9 & 105.8 & 76.9 & 288.2 & 7.2 & 84.0 \\
        \textit{TTO} & 209.6 & 261.8 & 59.7 & 105.9 & 139.6 & 343.0 & 6.5 & 64.2 \\
         \textbf{\textit{VC+TTO}} & \textbf{53.3} & \textbf{174.5} & \textbf{51.1} & \textbf{90.6} & \textbf{54.1} & \textbf{253.2} & \textbf{5.8} & \textbf{36.9} \\

        \bottomrule
    \end{tabularx}
\end{table}
}

{
\setlength{\tabcolsep}{0.0pt}

\begin{table}[t]
    \centering
    \small
    \caption{\textbf{Impact of auxiliary pseudo-depth estimation \textit{D\textsubscript{tr}}.}}
    \label{tab:ablation_aux}
    \begin{tabularx}{\linewidth}{l>{\centering}X>{\centering}X>{\centering}X>{\centering}X>{\centering}X>{\centering}X>{\centering}X>{\centering\arraybackslash}X}
        \toprule
        & \multicolumn{2}{c}{{MPJPE} $\downarrow$} & \multicolumn{2}{c}{{MPJPE-RA} $\downarrow$} & \multicolumn{2}{c}{{MRRPE} $\downarrow$} & \multicolumn{2}{c}{{L2} $\downarrow$} \\
        \cmidrule(lr){2-3}
        \cmidrule(lr){4-5}
        \cmidrule(lr){6-7}
        \cmidrule(lr){8-9}
        & H2O & AsHa & H2O & AsHa & H2O & AsHa & H2O & AsHa \\
        \midrule
        
        \textit{V-HPOT - D\textsubscript{tr}} & 60.6 & 237.7 & 56.1 & 98.1 & 78.9 & 310.1 & 5.9 & 56.1 \\
        \textit{V-HPOT(DA)} & 67.2 & 200.9 & \textbf{44.1} & \textbf{86.1} & 77.8 & \textbf{158.6} & \textbf{5.6} & 43.9 \\
        \textbf{\textit{V-HPOT}} & \textbf{53.3} & \textbf{174.5} & 51.1 & 90.6 & \textbf{54.1} & 253.2 & 5.8 & \textbf{36.9} \\
        
        \bottomrule
    \end{tabularx}
    
\end{table}
}

\vspace{-0.4cm}
\paragraph{Test-Time Optimisation Loss.}

We evaluated our TTO loss $\mathcal{L}_{\text{test-time}}$ through ablation studies: 
\begin{enumerate}

\item \textit{base}: No test-time optimisation
\item $\mathcal{L}_{\text{xy}}$: Consistency loss between $\text{P}_{L,R}^{\text{init}}$ and augmented pose $ \text{P}_{L,R}^{\rightarrow S_1}$ on 2D coordinates only, enforcing keypoint alignment in image space without considering depth.
\item $\mathcal{L}_{\text{z}}$: 
Consistency loss on $z$ only between $\text{P}_{L,R}^{\text{init}}, \text{P}_{L,R}^{\rightarrow S_1}$
\item $\mathcal{L}_{\text{xyz}}$: Combined consistency across all $x,y,z$ coordinates between $\text{P}_{L,R}^{\text{init}}, \text{P}_{L,R}^{\rightarrow S_1}$ poses
\item $\mathcal{L}_{\text{xyz}}^{n=2}$: We extend the consistency loss from one augmented pose ($n=1$) to two augmented poses ($n=2$) (\( \text{P}_{L,R}^{\text{init}}, \text{P}_{L,R}^{\rightarrow S_1}, \text{P}_{L,R}^{\rightarrow S_2} \))
\item $\mathcal{L}_{\text{xyz}}^{n=2}\!\!+\!\!D_{t}$: Adding pseudo-depth estimation to 5
\end{enumerate}

Findings in Table~\ref{tab:ablation_loss}. indicate that incorporating \( \mathcal{L}_{\text{xy}} \) improves 2D estimations in both target datasets. However, while the 2D error decreases, MPJPE increases in \textit{H2O}~\cite{Kwon_2021_ICCV} and decreases in \textit{AssemblyHands}~\cite{ohkawa:cvpr23}, suggesting that 2D improvements alone do not guarantee better 3D estimation. Applying \( \mathcal{L}_{\text{z}} \) results in lower MPJPE and 2D L2 error across both datasets. Combining both losses in \( \mathcal{L}_{\text{xyz}}\) further improves MPJPE and L2 error in \textit{H2O}~\cite{Kwon_2021_ICCV} while also outperforming the \textit{base} model in \textit{AssemblyHands}~\cite{ohkawa:cvpr23}. 

The $\mathcal{L}_{\text{xyz}}^{n=2}$ variant achieves the best results on both datasets, demonstrating the effectiveness of our virtual camera approach with multi-view consistency enforcement. Notably, incorporating pseudo-depth estimation at test time ($\mathcal{L}_{\text{xyz}}^{n=2}\!\!+\!\!D_{t}$) slightly increases the pose error, leading us to exclude it from the TTO process and use it only for training.

{
\setlength{\tabcolsep}{0.5pt}
\begin{table}[t]
    \centering
    \small
   \caption{\textbf{TTO strategies impact.} Proposed depth-augmented 3D pose consistency loss achieves best results.}
    \label{tab:ablation_loss}
    \begin{tabularx}{\linewidth}{l>{\centering}X>{\centering}X>{\centering}X>{\centering}X>{\centering}X>{\centering}X>{\centering}X>{\centering\arraybackslash}X}
        \toprule
        & \multicolumn{2}{c}{{MPJPE} $\downarrow$} & \multicolumn{2}{c}{{MPJPE-RA} $\downarrow$} & \multicolumn{2}{c}{{MRRPE} $\downarrow$} & \multicolumn{2}{c}{{L2} $\downarrow$} \\
        \cmidrule(lr){2-3}
        \cmidrule(lr){4-5}
        \cmidrule(lr){6-7}
        \cmidrule(lr){8-9}
        & H2O & AsHa & H2O & AsHa & H2O & AsHa & H2O & AsHa \\
        \midrule
        \textit{base} & 179.6 & 297.7 & 52.7 & 105.3 & 126.2 & 301.8 & 7.4 & 80.1 \\
        $\mathcal{L}_{xy}$ & 226 & 177.5 & 79.4 & 93.3 & 95.88 & 271.2 & 6.2 & 40.1 \\
        $\mathcal{L}_{z}$ & 54.7 & 253.2 & \textbf{49.1} & 124.5 & 62.5 & 338.7 & 6.1 & 64.7 \\
        $\mathcal{L}_{xyz}$ & 53.99 & 190.9 & 50.9 & 94.4 & 57.4 & 253.9 & 5.9 & 43.6 \\
        
        $\mathcal{L}_{xyz}^{n=2}$ & \textbf{53.3} & \textbf{174.5} & 51.1 & \textbf{90.6} & \textbf{54.1} & \textbf{253.2} & \textbf{5.8} & 36.9 \\

        $\mathcal{L}_{xyz}^{n=2}\!\!+\!\!D_{t}$ & 56.5 & 175.3 & 52.7 & 91.5 & 54.2 & 254.7 & 5.9 & \textbf{36.7} \\
        
        \bottomrule
    \end{tabularx}
\end{table}
}

\vspace{-0.4cm}
\paragraph{Comparison with other TTO approaches.}

As the fundamental difference between TTO methods for hand pose estimation lies in the self-supervision task, we create a unified TTO process and compare V-HPOT to \cite{spurr2021self}, introducing 2D consistency loss, and \cite{hu2024fast} with image inpainting to \textit{base} with excluded \textit{D\textsubscript{tr}}. Our comparison excludes methods \cite{cao2021reconstructing,hasson2020leveraging} which require object information or ground truth, or multi-view input~\cite{li2021test}. V-HPOT uniquely targets 3D understanding, achieving reductions in both 2D and 3D errors, whilst competing methods only show 2D gains (Table~\ref{tab:ablation_data}). While \cite{spurr2021self} also relies on consistency loss, it is limited to 2D keypoint transformations. V-HPOT's novelty lies in enforci ng 3D spatial consistency, making it the first method to effectively handle single-camera 3D hand pose estimation with superior performance over~\cite{spurr2021self,hu2024fast}.

\paragraph{Comparison with foundation model DINOv2.} In Table~\ref{tab:foundation} we compare DINOv2~\cite{oquab2023dinov2} cross-domain generalising capabilities with frozen weights \textit{(D2)}, fine-tuned last 4 layers \textit{(D2-FT)}, and as V-HPOT backbone \textit{(V-HPOT-D2)}. Results show \textit{D2} and \textit{D2-FT} cross-domain generalisation is insufficient in our task. As backbone, performance is comparable but inferior to \textit{EfficientNetv2s} while being larger (86M vs. 21M parameters) and inferring $\approx$5× slower (109.1 vs. 22.8ms mean of 1000 trials per batch on RTX3090).

\subsection{Comparison with the State of the Art}

We compare our approach to other egocentric 3D hand pose estimation methods in a cross-domain setting, using MPJPE-RA for 3D pose accuracy and MRRPE for absolute 3D error, following~\cite{wildhands2025}. Our method outperforms other single-stage approaches, including H2OTR~\cite{cho2023transformer} and ArcticNet-SF~\cite{wildhands2025}, in every scenario, including in-the-wild Epic-Kpts~\cite{wildhands2025}, despite being trained on less data (Table~\ref{tab:sota_comparison}).

Direct comparisons between V-HPOT and two-stage methods disadvantage our approach for three critical reasons: (1) two-stage methods utilise hand detectors trained on massive, diverse datasets while we rely solely on limited egocentric data; (2) they process high-resolution hand crops, whereas we handle entire scenes at 224×224 pixels with hands occupying only a minimal area; and (3) they leverage larger training datasets.
Despite these disadvantages, V-HPOT outperforms FrankMocap across all datasets while being trained on $\approx\!\!\times4$ less data. We surpass HaMer on two datasets in MRRPE (absolute 3D error) with $\approx\!\!\times\!14$ less training data and trail WildHands by only 5mm on H2O, which uses $\approx\!\!\times 3.5$ more training data. Most notably, our 2D L2 error is merely 1.2 pixels behind WildHands in \textit{Epic-Kpts}, despite our whole-scene approach versus their focused hand crops, showing the strength of VC and our TTO in improving cross-domain hand pose performance.
{
\setlength{\tabcolsep}{0.5pt}
\begin{table}[t]
\centering
\small
\caption{\textbf{Comparison of different} TTO/self-supervised strategies.}
\label{tab:ablation_data}
\begin{tabularx}{\linewidth}{l>{\centering}X>{\centering}X>{\centering}X>{\centering}X>{\centering}X>{\centering}X>{\centering}X>{\centering\arraybackslash}X}
    \toprule
    & \multicolumn{2}{c}{{MPJPE} $\downarrow$} & \multicolumn{2}{c}{{MPJPE-RA} $\downarrow$} & \multicolumn{2}{c}{{MRRPE} $\downarrow$} & \multicolumn{2}{c}{{L2} $\downarrow$} \\
    \cmidrule(lr){2-3}
    \cmidrule(lr){4-5}
    \cmidrule(lr){6-7}
    \cmidrule(lr){8-9}
    & H2O & AsHa & H2O & AsHa & H2O & AsHa & H2O & AsHa \\
    \midrule
    \textit{base} & 139.1 & 329.2 & 53.2 & 112.6 & 111.1 & 463.3 & 8.0 & 88.3 \\
    \textit{PeCLR}~\cite{spurr2021self} & 106.1 & 384.3 & 56.0 & 111.5 & 95.9 & 448.1 & 7.4 & 57.8 \\
    \textit{MeTTA}~\cite{hu2024fast} & 142.2 & 349.9 & 53.3 & 108.5 & 112.3 & 413.9 & 7.8 & 86.1 \\
    \textbf{\textit{V-HPOT}} & \textbf{53.3} & \textbf{174.5} & \textbf{51.1} & \textbf{90.6} & \textbf{54.1} & \textbf{253.2} & \textbf{5.8} & \textbf{36.9} \\
    \bottomrule
\end{tabularx}
\end{table}
}

{
\setlength{\tabcolsep}{0.5pt}
\begin
{table}[t]
\centering
\small
\caption{\textbf{Foundation model DINOv2} as a backbone in frozen and fine-tuned setup, and with our V-HPOT TTO pipeline.}
\label{tab:foundation}
\begin{tabularx}{\linewidth}{l>{\centering}X>{\centering}X>{\centering}X>{\centering}X>{\centering}X>{\centering}X>{\centering}X>{\centering\arraybackslash}X}
    \toprule
    & \multicolumn{2}{c}{{MPJPE} $\downarrow$} & \multicolumn{2}{c}{{MPJPE-RA} $\downarrow$} & \multicolumn{2}{c}{{MRRPE} $\downarrow$} & \multicolumn{2}{c}{{L2} $\downarrow$} \\
    \cmidrule(lr){2-3}
    \cmidrule(lr){4-5}
    \cmidrule(lr){6-7}
    \cmidrule(lr){8-9}
    & H2O & AsHa & H2O & AsHa & H2O & AsHa & H2O & AsHa \\
    \midrule
    \textit{D2}  & 226.7 & 259.7 & 78.8 & 151.1 & 160.7 & 468.6 & 32.8 & 48.7 \\
    \textit{D2-FT} & 185.7 & 249.8 & 58.1 & 132.1 & 127.8 & 426.2 & 7.9 & 45.8 \\
    \textit{V-HPOT-D2} & 59.4 & 191.2 & 58.0 & 93.2 & 64.1 & \textbf{221.8} & 7.5 & 44.3 \\
    \textbf{\textit{V-HPOT}} & \textbf{53.3} & \textbf{174.5} & \textbf{51.1} & \textbf{90.6} & \textbf{54.1} & 253.2 & \textbf{5.8} & \textbf{36.9} \\
    \bottomrule
\end{tabularx}
\vspace{-0.5cm}
\end{table}
}

{
\setlength{\tabcolsep}{0.5pt}
\begin{table*}[h!]
    \centering
    \caption{\textbf{Comparison of state-of-the-art egocentric hand pose models in cross-domain settings.} The table indicates whether a model follows a single-stage approach, the number of training domains and images, where the number in brackets for \cite{wildhands2025,fan2023arctic} relates to the model tested on \textit{AssemblyHands}. MPJPE-RA and MRRPE are reported in millimetres. *  means results are referenced from~\cite{wildhands2025}.}
    \label{tab:sota_comparison}
    \renewcommand{\arraystretch}{0.9}
    \begin{tabularx}{\linewidth}{l>{\centering}X>{\centering}X>{\centering}X>{\centering}X>{\centering}X>{\centering}X>{\centering}X>{\centering\arraybackslash}X}
        \toprule
        & & & & \multicolumn{2}{c}{\textbf{H2O}} & \multicolumn{2}{c}{\textbf{Assembly}} & \textbf{Epic-Kps} \\
        \cmidrule(lr){5-6}
        \cmidrule(lr){7-8}
        \cmidrule(lr){9-9}
        {Method} & {Single Stage} & {T. Domains} & {T. Img} & MPJPE-RA & MRRPE & MPJPE-RA & MRRPE & L2 \\
        \midrule
        \rowcolor{gray!20}FrankMocap \cite{rong2021frankmocap} & \ding{55} & 6 & 675K & 58.51 & - & 97.59 & - & 13.33 \\
        \rowcolor{gray!20}HaMeR \cite{hamerpavlakos2024reconstructing} & \ding{55} & 11 & 2.7M & \textbf{23.82} & 147.87 & \textbf{45.49} & 334.52 & \textbf{4.56} \\
        \rowcolor{gray!20}WildHands \cite{wildhands2025} & \ding{55} & 3 & 627K(267K) & 31.08 & 49.49 & 80.40 & \textbf{148.12} & 7.20 \\
        \midrule
        \midrule
        ArcticNet-SF \cite{fan2023arctic}* & \checkmark & 3 & 627K(267K) & 83.84 & 325.55 & 110.76 & 326.94 & 35.02 \\
        H2OTR~\cite{cho2023transformer} & \checkmark & 1 & 55K & 115.16 & 418.59 & 137.25 & 543.28 & 14.46 \\
        SHARP~\cite{mucha2025sharp} & \checkmark & 1 & 180K & 52.40 & 98.08 & 103.84 & 282.21 & 14.41 \\
        
        \textbf{V-HPOT (Ours)} & \checkmark & \textbf{1} & 180K & \textbf{51.11} & \textbf{54.07} & \textbf{92.09} & \textbf{252.85} & \textbf{8.45} \\
        \bottomrule
    \end{tabularx}
\end{table*}
}




\subsection{Qualitative Results}

\begin{figure*}[th]
  \centering
  \includegraphics[width=\linewidth]{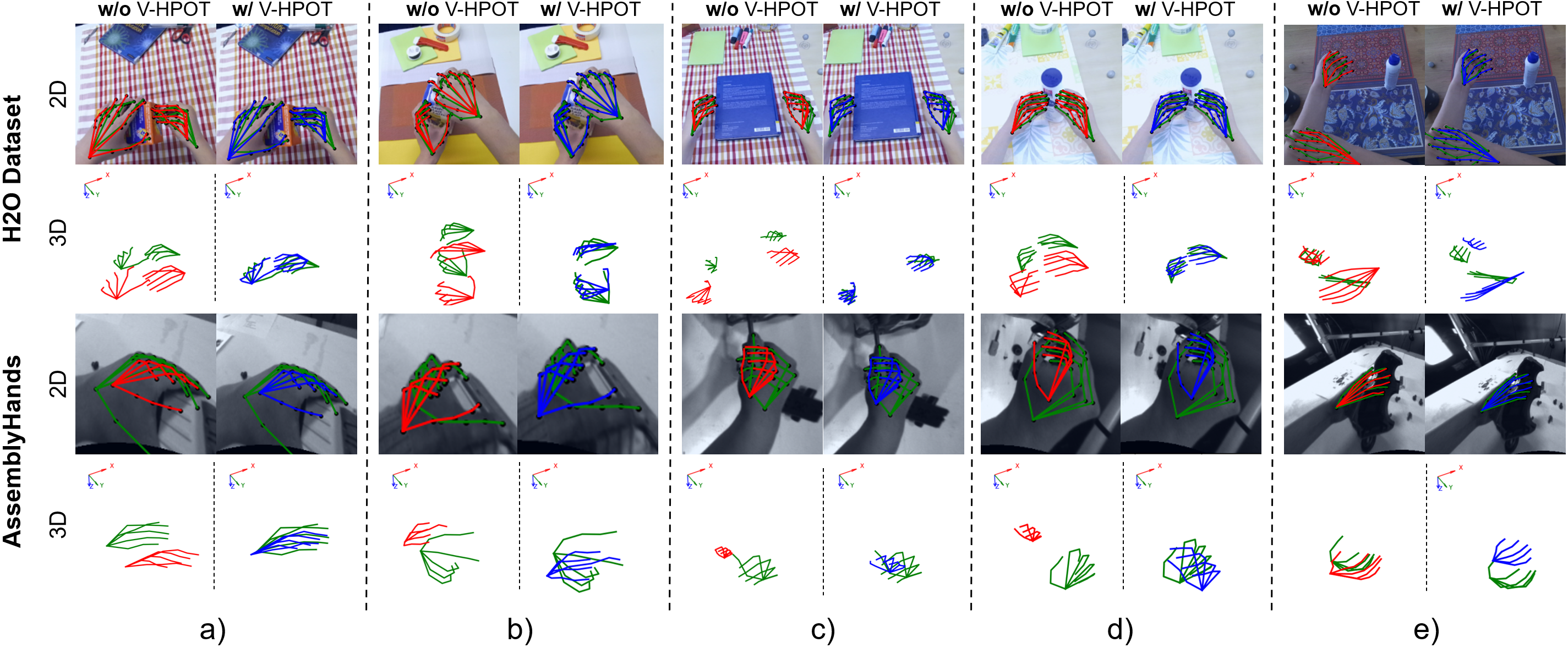}
  \caption[Caption for LOF]{
\textbf{Qualitative results of our method in 2D and 3D space.} Green skeletons represent the \textcolor{ForestGreen}{ground truth hand pose}, red estimations \textcolor{red}{\textbf{without} \textit{V-HPOT}} and blue estimations \textcolor{blue}{\textbf{with} \textit{V-HPOT}}. Four examples from the left (a-d) show that \textit{V-HPOT} improves 3D pose estimation, while two examples from the right side (e) present a negative case with an increase of error.
  
  }
  \label{fig:qualitative}
  \vspace{-0.4cm}
\end{figure*}

We present qualitative comparisons with and without V-HPOT in Figure~\ref{fig:qualitative}, showing 2D pose projections and corresponding 3D poses. Images (a–d) demonstrate V-HPOT improvements, while (e) shows a negative impact of V-HPOT. In \textit{H2O Dataset}~\cite{Kwon_2021_ICCV}, quantitative errors are lower than in \textit{AssemblyHands}~\cite{ohkawa:cvpr23}, which is also reflected in the qualitative results. In \textit{H2O}, V-HPOT performs well in scenarios where the subject interacts with objects, with the most significant improvements observed in wrist positioning. We identify failure cases in both datasets when hands move close to or directly towards the camera (Figure~\ref{fig:qualitative}-e). Such close-camera scenarios are outliers in typical egocentric datasets, which focus on object manipulation. Therefore, improvements to data collection are required.
Additionally, \textit{AssemblyHands} contains stronger distortions, where V-HPOT tends to fail more frequently (Figure~\ref{fig:qualitative}-e), which could be addressed with dedicated distortion-aware augmentations or normalising field of view across datasets.

\section{Conclusion}

\label{sec:conclusion}

We presented V-HPOT, a novel approach to egocentric 3D hand pose estimation that addresses the challenge of cross-domain generalisation. Our method introduces two key innovations: (1) a virtual camera transformation that decouples pose estimation from camera-specific intrinsics, enhancing depth representation learning; and (2) a self-supervised TTO framework that leverages consistency constraints between depth-augmented predictions.
Our TTO strategy -- the first to specifically target depth understanding in hand pose estimation -- performs continuous weight updates during inference on the initial samples of target data. Comparison with the foundation model, particularly DINOv2, justifies the need for dedicated cross-domain hand pose studies like V-HPOT.
V-HPOT's effectiveness is confirmed through improvements in 3D absolute error (MPJPE and MRRPE) across datasets, reducing MPJPE by 70\% on \textit{H2O}, 41\% on \textit{AssemblyHands}, despite the latter's domain shift encompassing different scenes, sensors, and image modality (monochromatic vs. RGB).
V-HPOT outperforms state-of-the-art single-stage approaches in cross-domain egocentric hand pose estimation while remaining competitive with two-stage methods despite using $\approx\times3.5- \times14$ fewer training samples.
The model demonstrates strong performance on real-world benchmark \textit{Epic-Kpts} with 30\% L2 error improvements, which is important as most 3D hand pose datasets are recorded in lab environments due to labelling procedures. Qualitative analysis confirms that V-HPOT produces anatomically plausible hand estimates even during complex object manipulation tasks. All this is achieved without relying on additional large-scale pretraining, datasets, or labels, confirming V-HPOT as a step forward in understanding cross-domain single-camera egocentric 3D hand pose.

\paragraph{Acknowledgments.} This research was supported by VisuAAL ITN H2020 (grant agreement no. 861091) and by the iMars project MSCA (grant agreement no.101182996).

{
    \small
    \bibliographystyle{ieeenat_fullname}
    \bibliography{main}

\begin{thebibliography}{45}
\providecommand{\natexlab}[1]{#1}
\providecommand{\url}[1]{\texttt{#1}}
\expandafter\ifx\csname urlstyle\endcsname\relax
  \providecommand{\doi}[1]{doi: #1}\else
  \providecommand{\doi}{doi: \begingroup \urlstyle{rm}\Url}\fi

\bibitem[Aboukhadra et~al.(2023)Aboukhadra, Malik, Elhayek, Robertini, and Stricker]{aboukhadra2023thor-net}
Ahmed Aboukhadra, Jameel Malik, Ahmed Elhayek, Nadia Robertini, and Didier Stricker.
\newblock {THOR-Net: End-to-end Graformer-based Realistic Two Hands and Object Reconstruction with Self-supervision}.
\newblock In \emph{Proceedings of the IEEE/CVF Winter Conference on Applications of Computer Vision}, pages 1001--1010, 2023.

\bibitem[Banerjee et~al.(2025)Banerjee, Shkodrani, Moulon, Hampali, Han, Zhang, Zhang, Fountain, Miller, Basol, et~al.]{banerjee2024hot3d}
Prithviraj Banerjee, Sindi Shkodrani, Pierre Moulon, Shreyas Hampali, Shangchen Han, Fan Zhang, Linguang Zhang, Jade Fountain, Edward Miller, Selen Basol, et~al.
\newblock {HOT3D: Hand and Object Tracking in 3D from Egocentric Multi-View Videos}.
\newblock In \emph{Proceedings of the Computer Vision and Pattern Recognition Conference}, pages 7061--7071, 2025.

\bibitem[Brazil et~al.(2023)Brazil, Kumar, Straub, Ravi, Johnson, and Gkioxari]{brazil2023omni3d}
Garrick Brazil, Abhinav Kumar, Julian Straub, Nikhila Ravi, Justin Johnson, and Georgia Gkioxari.
\newblock {OMNI3D: A Large Benchmark and Model for 3D Object Detection in the Wild}.
\newblock In \emph{Proceedings of the IEEE/CVF Conference on Computer Vision and Pattern Recognition}, pages 13154--13164, 2023.

\bibitem[Cao et~al.(2021)Cao, Radosavovic, Kanazawa, and Malik]{cao2021reconstructing}
Zhe Cao, Ilija Radosavovic, Angjoo Kanazawa, and Jitendra Malik.
\newblock {Reconstructing Hand-Object Interactions in the Wild}.
\newblock In \emph{Proceedings of the IEEE/CVF International Conference on Computer Vision}, pages 12417--12426, 2021.

\bibitem[Chen et~al.(2021{\natexlab{a}})Chen, Lin, Xie, Lin, and Xie]{chen2021temporal}
Liangjian Chen, Shih-Yao Lin, Yusheng Xie, Yen-Yu Lin, and Xiaohui Xie.
\newblock {Temporal-Aware Self-Supervised Learning for 3D Hand Pose and Mesh Estimation in Videos}.
\newblock In \emph{Proceedings of the IEEE/CVF Winter Conference on Applications of Computer Vision}, pages 1050--1059, 2021{\natexlab{a}}.

\bibitem[Chen et~al.(2021{\natexlab{b}})Chen, Tu, Kang, Bao, Zhang, Zhe, Chen, and Yuan]{chen2021model}
Yujin Chen, Zhigang Tu, Di Kang, Linchao Bao, Ying Zhang, Xuefei Zhe, Ruizhi Chen, and Junsong Yuan.
\newblock {Model-based 3D Hand Reconstruction via Self-Supervised Learning}.
\newblock In \emph{Proceedings of the IEEE/CVF Conference on Computer Vision and Pattern Recognition}, pages 10451--10460, 2021{\natexlab{b}}.

\bibitem[Cho et~al.(2023)Cho, Kim, Kim, Lee, Ismayilzada, and Baek]{cho2023transformer}
Hoseong Cho, Chanwoo Kim, Jihyeon Kim, Seongyeong Lee, Elkhan Ismayilzada, and Seungryul Baek.
\newblock {Transformer-Based Unified Recognition of Two Hands Manipulating Objects}.
\newblock In \emph{Proceedings of the IEEE/CVF Conference on Computer Vision and Pattern Recognition}, pages 4769--4778, 2023.

\bibitem[Damen et~al.(2018)Damen, Doughty, Farinella, Fidler, Furnari, Kazakos, Moltisanti, Munro, Perrett, Price, and Wray]{Damen2018EPICKITCHENS}
Dima Damen, Hazel Doughty, Giovanni~Maria Farinella, Sanja Fidler, Antonino Furnari, Evangelos Kazakos, Davide Moltisanti, Jonathan Munro, Toby Perrett, Will Price, and Michael Wray.
\newblock {Scaling Egocentric Vision: The EPIC-KITCHENS Dataset}.
\newblock In \emph{European Conference on Computer Vision (ECCV)}, 2018.

\bibitem[Darkhalil et~al.(2022)Darkhalil, Shan, Zhu, Ma, Kar, Higgins, Fidler, Fouhey, and Damen]{VISOR2022}
Ahmad Darkhalil, Dandan Shan, Bin Zhu, Jian Ma, Amlan Kar, Richard Higgins, Sanja Fidler, David Fouhey, and Dima Damen.
\newblock {EPIC-KITCHENS VISOR Benchmark: VIdeo Segmentations and Object Relations}.
\newblock In \emph{Proceedings of the Neural Information Processing Systems (NeurIPS) Track on Datasets and Benchmarks}, 2022.

\bibitem[Dosovitskiy et~al.(2021)Dosovitskiy, Beyer, Kolesnikov, Weissenborn, Zhai, Unterthiner, Dehghani, Minderer, Heigold, Gelly, Uszkoreit, and Houlsby]{dosovitskiy2020image}
Alexey Dosovitskiy, Lucas Beyer, Alexander Kolesnikov, Dirk Weissenborn, Xiaohua Zhai, Thomas Unterthiner, Mostafa Dehghani, Matthias Minderer, Georg Heigold, Sylvain Gelly, Jakob Uszkoreit, and Neil Houlsby.
\newblock {An Image is Worth 16x16 Words: Transformers for Image Recognition at Scale}.
\newblock In \emph{International Conference on Learning Representations}, 2021.

\bibitem[Fan et~al.(2023)Fan, Taheri, Tzionas, Kocabas, Kaufmann, Black, and Hilliges]{fan2023arctic}
Zicong Fan, Omid Taheri, Dimitrios Tzionas, Muhammed Kocabas, Manuel Kaufmann, Michael~J Black, and Otmar Hilliges.
\newblock {ARCTIC: A Dataset for Dexterous Bimanual Hand-Object Manipulation}.
\newblock In \emph{Proceedings of the IEEE/CVF Conference on Computer Vision and Pattern Recognition}, pages 12943--12954, 2023.

\bibitem[Fang et~al.(2024)Fang, Chen, Wang, Zhang, Zhang, Xu, He, Gao, Su, Li, et~al.]{fang2024egopat3dv2}
Irving Fang, Yuzhong Chen, Yifan Wang, Jianghan Zhang, Qiushi Zhang, Jiali Xu, Xibo He, Weibo Gao, Hao Su, Yiming Li, et~al.
\newblock {EgoPAT3Dv2: Predicting 3D Action Target from 2D Egocentric Vision for Human-Robot Interaction}.
\newblock In \emph{2024 IEEE International Conference on Robotics and Automation (ICRA)}, pages 3036--3043. IEEE, 2024.

\bibitem[Grauman et~al.(2022)Grauman, Westbury, Byrne, Chavis, Furnari, Girdhar, Hamburger, Jiang, Liu, Liu, et~al.]{grauman2022ego4d}
Kristen Grauman, Andrew Westbury, Eugene Byrne, Zachary Chavis, Antonino Furnari, Rohit Girdhar, Jackson Hamburger, Hao Jiang, Miao Liu, Xingyu Liu, et~al.
\newblock {Ego4D: Around the World in 3,000 Hours of Egocentric Video}.
\newblock In \emph{Proceedings of the IEEE/CVF Conference on Computer Vision and Pattern Recognition}, pages 18995--19012, 2022.

\bibitem[Hasson et~al.(2019)Hasson, Varol, Tzionas, Kalevatykh, Black, Laptev, and Schmid]{hasson2019learning}
Yana Hasson, Gul Varol, Dimitrios Tzionas, Igor Kalevatykh, Michael~J Black, Ivan Laptev, and Cordelia Schmid.
\newblock Learning joint reconstruction of hands and manipulated objects.
\newblock In \emph{Proceedings of the IEEE/CVF Conference on Computer Vision and Pattern Recognition}, pages 11807--11816, 2019.

\bibitem[Hasson et~al.(2020)Hasson, Tekin, Bogo, Laptev, Pollefeys, and Schmid]{hasson2020leveraging}
Yana Hasson, Bugra Tekin, Federica Bogo, Ivan Laptev, Marc Pollefeys, and Cordelia Schmid.
\newblock {Leveraging Photometric Consistency over Time for Sparsely Supervised Hand-Object Reconstruction}.
\newblock In \emph{Proceedings of the IEEE/CVF Conference on Computer Vision and Pattern Recognition}, pages 571--580, 2020.

\bibitem[Hu et~al.(2024)Hu, Sun, Li, Wei, Li, and Lu]{hu2024fast}
Shengxiang Hu, Huaijiang Sun, Bin Li, Dong Wei, Weiqing Li, and Jianfeng Lu.
\newblock {Fast Adaptation for Human Pose Estimation via Meta-Optimization}.
\newblock In \emph{Proceedings of the IEEE/CVF Conference on Computer Vision and Pattern Recognition}, pages 1792--1801, 2024.

\bibitem[Kanazawa et~al.(2018)Kanazawa, Tulsiani, Efros, and Malik]{kanazawa2018learning}
Angjoo Kanazawa, Shubham Tulsiani, Alexei~A Efros, and Jitendra Malik.
\newblock {Learning Category-Specific Mesh Reconstruction from Image Collections}.
\newblock In \emph{Proceedings of the European Conference on Computer Vision (ECCV)}, pages 371--386, 2018.

\bibitem[Karunratanakul et~al.(2020)Karunratanakul, Yang, Zhang, Black, Muandet, and Tang]{karunratanakul2020grasping}
Korrawe Karunratanakul, Jinlong Yang, Yan Zhang, Michael~J Black, Krikamol Muandet, and Siyu Tang.
\newblock {Grasping Field: Learning Implicit Representations for Human Grasps}.
\newblock In \emph{2020 International Conference on 3D Vision (3DV)}, pages 333--344. IEEE, 2020.

\bibitem[Kwon et~al.(2021)Kwon, Tekin, St\"uhmer, Bogo, and Pollefeys]{Kwon_2021_ICCV}
Taein Kwon, Bugra Tekin, Jan St\"uhmer, Federica Bogo, and Marc Pollefeys.
\newblock {H2O: Two Hands Manipulating Objects for First Person Interaction Recognition}.
\newblock In \emph{Proceedings of the IEEE/CVF International Conference on Computer Vision (ICCV)}, pages 10138--10148, 2021.

\bibitem[Li et~al.(2016)Li, Wang, Shi, Liu, and Hou]{li2016revisiting}
Yanghao Li, Naiyan Wang, Jianping Shi, Jiaying Liu, and Xiaodi Hou.
\newblock {Revisiting Batch Normalization For Practical Domain Adaptation}.
\newblock \emph{arXiv preprint arXiv:1603.04779}, 2016.

\bibitem[Li et~al.(2021)Li, Hao, Di, Gundavarapu, and Wang]{li2021test}
Yizhuo Li, Miao Hao, Zonglin Di, Nitesh~Bharadwaj Gundavarapu, and Xiaolong Wang.
\newblock {Test-Time Personalization with a Transformer for Human Pose Estimation}.
\newblock \emph{Advances in Neural Information Processing Systems}, 34:\penalty0 2583--2597, 2021.

\bibitem[Lin et~al.(2025)Lin, Ohkawa, Huang, Zhang, Cai, Li, Furuta, and Sato]{lin2025simhand}
Nie Lin, Takehiko Ohkawa, Yifei Huang, Mingfang Zhang, Minjie Cai, Ming Li, Ryosuke Furuta, and Yoichi Sato.
\newblock {Si{MH}and: Mining Similar Hands for Large-Scale 3D Hand Pose Pre-training}.
\newblock In \emph{The Thirteenth International Conference on Learning Representations}, 2025.

\bibitem[Lin et~al.(2017)Lin, Doll{\'a}r, Girshick, He, Hariharan, and Belongie]{lin2017feature}
Tsung-Yi Lin, Piotr Doll{\'a}r, Ross Girshick, Kaiming He, Bharath Hariharan, and Serge Belongie.
\newblock {Feature Pyramid Networks for Object Detection}.
\newblock In \emph{Proceedings of the IEEE Conference on Computer Vision and Pattern Recognition}, pages 2117--2125, 2017.

\bibitem[Liu et~al.(2021)Liu, Jiang, Xu, Liu, and Wang]{liu2021semi}
Shaowei Liu, Hanwen Jiang, Jiarui Xu, Sifei Liu, and Xiaolong Wang.
\newblock {Semi-Supervised 3D Hand-Object Poses Estimation with Interactions in Time}.
\newblock In \emph{Proceedings of the IEEE/CVF Conference on Computer Vision and Pattern Recognition}, pages 14687--14697, 2021.

\bibitem[Moon et~al.(2020)Moon, Yu, Wen, Shiratori, and Lee]{moon2020interhand2}
Gyeongsik Moon, Shoou-I Yu, He Wen, Takaaki Shiratori, and Kyoung~Mu Lee.
\newblock {InterHand2.6M: A Dataset and Baseline for 3D Interacting Hand Pose Estimation from a Single RGB Image}.
\newblock In \emph{European Conference on Computer Vision}, pages 548--564. Springer, 2020.

\bibitem[Mucha and Kampel(2024)]{mucha2024my}
Wiktor Mucha and Martin Kampel.
\newblock {In My Perspective, in My Hands: Accurate Egocentric 2D Hand Pose and Action Recognition}.
\newblock In \emph{2024 IEEE 18th International Conference on Automatic Face and Gesture Recognition (FG)}, pages 1--9, 2024.

\bibitem[Mucha et~al.(2024)Mucha, Cuconasu, Etori, Kalokyri, and Trappolini]{text2taste}
Wiktor Mucha, Florin Cuconasu, Naome~A. Etori, Valia Kalokyri, and Giovanni Trappolini.
\newblock {TEXT2TASTE: A Versatile Egocentric Vision System for Intelligent Reading Assistance Using Large Language Model}.
\newblock In \emph{Computers Helping People with Special Needs}, pages 285--291, Cham, 2024. Springer Nature Switzerland.

\bibitem[Mucha et~al.(2025)Mucha, Wray, and Kampel]{mucha2025sharp}
Wiktor Mucha, Michael Wray, and Martin Kampel.
\newblock {SHARP: Segmentation of Hands and Arms by Range Using Pseudo-depth for Enhanced Egocentric 3D Hand Pose Estimation and Action Recognition}.
\newblock In \emph{International Conference on Pattern Recognition}, pages 178--193. Springer, 2025.

\bibitem[Ohkawa et~al.(2023)Ohkawa, He, Sener, Hodan, Tran, and Keskin]{ohkawa:cvpr23}
Takehiko Ohkawa, Kun He, Fadime Sener, Tomas Hodan, Luan Tran, and Cem Keskin.
\newblock {{AssemblyHands:} Towards Egocentric Activity Understanding via 3D Hand Pose Estimation}.
\newblock In \emph{Proceedings of the IEEE/CVF Conference on Computer Vision and Pattern Recognition (CVPR)}, pages 12999--13008, 2023.

\bibitem[Oquab et~al.(2024)Oquab, Darcet, Moutakanni, Vo, Szafraniec, Khalidov, Fernandez, Haziza, Massa, El-Nouby, et~al.]{oquab2023dinov2}
Maxime Oquab, Timoth{\'e}e Darcet, Th{\'e}o Moutakanni, Huy Vo, Marc Szafraniec, Vasil Khalidov, Pierre Fernandez, Daniel Haziza, Francisco Massa, Alaaeldin El-Nouby, et~al.
\newblock {DINOv2: Learning Robust Visual Features without Supervision}.
\newblock \emph{Transactions on Machine Learning Research}, 2024.

\bibitem[Park et~al.(2022)Park, Oh, Moon, Choi, and Lee]{park2022handoccnet}
JoonKyu Park, Yeonguk Oh, Gyeongsik Moon, Hongsuk Choi, and Kyoung~Mu Lee.
\newblock {HandOccNet: Occlusion-Robust 3D Hand Mesh Estimation Network}.
\newblock In \emph{Proceedings of the IEEE/CVF Conference on Computer Vision and Pattern Recognition}, pages 1496--1505, 2022.

\bibitem[Pavlakos et~al.(2024)Pavlakos, Shan, Radosavovic, Kanazawa, Fouhey, and Malik]{hamerpavlakos2024reconstructing}
Georgios Pavlakos, Dandan Shan, Ilija Radosavovic, Angjoo Kanazawa, David Fouhey, and Jitendra Malik.
\newblock {Reconstructing Hands in 3D with Transformers}.
\newblock In \emph{Proceedings of the IEEE/CVF Conference on Computer Vision and Pattern Recognition}, pages 9826--9836, 2024.

\bibitem[Peng et~al.(2023)Peng, Zheng, and Chen]{peng2023source}
Qucheng Peng, Ce Zheng, and Chen Chen.
\newblock {Source-free Domain Adaptive Human Pose Estimation}.
\newblock In \emph{Proceedings of the IEEE/CVF International Conference on Computer Vision}, pages 4826--4836, 2023.

\bibitem[Prakash et~al.(2025)Prakash, Tu, Chang, and Gupta]{wildhands2025}
Aditya Prakash, Ruisen Tu, Matthew Chang, and Saurabh Gupta.
\newblock {3D Hand Pose Estimation in Everyday Egocentric Images}.
\newblock In \emph{European Conference on Computer Vision}, pages 183--202. Springer, 2025.

\bibitem[Ranftl et~al.(2021)Ranftl, Bochkovskiy, and Koltun]{ranftl2021vision}
Ren{\'e} Ranftl, Alexey Bochkovskiy, and Vladlen Koltun.
\newblock {Vision Transformers for Dense Prediction}.
\newblock In \emph{{Proceedings of the IEEE/CVF International Conference on Computer Vision}}, pages 12179--12188, 2021.

\bibitem[Rong et~al.(2021)Rong, Shiratori, and Joo]{rong2021frankmocap}
Yu Rong, Takaaki Shiratori, and Hanbyul Joo.
\newblock {FrankMocap: A Monocular 3D Whole-Body Pose Estimation System via Regression and Integration}.
\newblock In \emph{IEEE International Conference on Computer Vision Workshops}, 2021.

\bibitem[Spurr et~al.(2021)Spurr, Dahiya, Wang, Zhang, and Hilliges]{spurr2021self}
Adrian Spurr, Aneesh Dahiya, Xi Wang, Xucong Zhang, and Otmar Hilliges.
\newblock {Self-Supervised 3D Hand Pose Estimation from monocular RGB via Contrastive Learning}.
\newblock In \emph{Proceedings of the IEEE/CVF International Conference on Computer Vision}, pages 11230--11239, 2021.

\bibitem[Sun et~al.(2020)Sun, Wang, Liu, Miller, Efros, and Hardt]{sun2020test}
Yu Sun, Xiaolong Wang, Zhuang Liu, John Miller, Alexei Efros, and Moritz Hardt.
\newblock {Test-Time Training with Self-Supervision for Generalization under Distribution Shifts}.
\newblock In \emph{International Conference on Machine Learning}, pages 9229--9248. PMLR, 2020.

\bibitem[Tan and Le(2021)]{tan2021efficientnetv2}
Mingxing Tan and Quoc Le.
\newblock {EfficientNetV2: Smaller Models and Faster Training}.
\newblock In \emph{International Conference on Machine Learning}, pages 10096--10106. PMLR, 2021.

\bibitem[Tekin et~al.(2019)Tekin, Bogo, and Pollefeys]{tekin2019h+}
Bugra Tekin, Federica Bogo, and Marc Pollefeys.
\newblock {H+O: Unified Egocentric Recognition of 3D Hand-object Poses and Interactions}.
\newblock In \emph{Proceedings of the IEEE/CVF Conference on Computer Vision and Pattern Recognition}, pages 4511--4520, 2019.

\bibitem[Wen et~al.(2023)Wen, Pan, Yang, Pan, Komura, and Wang]{wen2023hierarchical}
Yilin Wen, Hao Pan, Lei Yang, Jia Pan, Taku Komura, and Wenping Wang.
\newblock {Hierarchical Temporal Transformer for 3D Hand Pose Estimation and Action Recognition from Egocentric RGB Videos}.
\newblock In \emph{Proceedings of the IEEE/CVF Conference on Computer Vision and Pattern Recognition}, pages 21243--21253, 2023.

\bibitem[Yang et~al.(2021)Yang, Chen, and Yao]{yang2021semihand}
Linlin Yang, Shicheng Chen, and Angela Yao.
\newblock {Semi-Supervised Hand Pose Estimation With Consistency}.
\newblock In \emph{Proceedings of the IEEE/CVF International Conference on Computer Vision}, pages 11364--11373, 2021.

\bibitem[Yang et~al.(2024)Yang, Kang, Huang, Xu, Feng, and Zhao]{yang2024depth}
Lihe Yang, Bingyi Kang, Zilong Huang, Xiaogang Xu, Jiashi Feng, and Hengshuang Zhao.
\newblock {Depth Anything: Unleashing the Power of Large-Scale Unlabeled Data}.
\newblock In \emph{Proceedings of the IEEE/CVF Conference on Computer Vision and Pattern Recognition}, pages 10371--10381, 2024.

\bibitem[Ye et~al.(2022)Ye, Gupta, and Tulsiani]{ye2022s}
Yufei Ye, Abhinav Gupta, and Shubham Tulsiani.
\newblock {What’s in your hands? 3D Reconstruction of Generic Objects in Hands}.
\newblock In \emph{Proceedings of the IEEE/CVF Conference on Computer Vision and Pattern Recognition}, pages 3895--3905, 2022.

\bibitem[Zheng et~al.(2023)Zheng, Wen, Xue, Ren, and Wang]{zheng2023hamuco}
Xiaozheng Zheng, Chao Wen, Zhou Xue, Pengfei Ren, and Jingyu Wang.
\newblock {HaMuCo: Hand Pose Estimation via Multiview Collaborative Self-Supervised Learning}.
\newblock In \emph{Proceedings of the IEEE/CVF International Conference on Computer Vision}, pages 20763--20773, 2023.

\end{thebibliography}
}

\clearpage
\twocolumn[
  \begin{@twocolumnfalse}
    \begin{center}
      {\Large\bfseries Supplementary Material: \\ Towards Egocentric 3D Hand Pose Estimation in Unseen Domains}
    \end{center}
    \vspace{1em}
  \end{@twocolumnfalse}
]

\appendix

\section{Additional Results and Experiments}
\subsection{Comparison of V-HPOT in H2O Dataset as source-target}

To evaluate our model's effectiveness relative to current state-of-the-art approaches, we conducted same-domain experiments using the \textit{H2O Dataset}~\cite{Kwon_2021_ICCV} for both training and testing. The results demonstrate that our model outperforms existing methods. The second-best performing approach, H2OTR~\cite{cho2023transformer}, employs a substantially more complex architecture incorporating hand-object contact maps. Yet, our more streamlined design still yields better results (See Table \ref{tab:h2o_res}). Our solution's strong performance is due to the 3D augmentation enabled by virtual camera space.


\begin{table}[t]
\small
\centering
\caption{\textbf{Results of 3D hand pose estimation} state-of-the-art methods when trained and tested within the same domain - \textit{H2O Dataset}. All numbers provided in \textit{mm} in camera space.}
\label{tab:h2o_res}
\begin{tabularx}{\columnwidth}{l>{\centering}X>{\centering\arraybackslash}X}
\toprule
Method       & Year    & MPJPE $\downarrow$ \\
\midrule
LPC \cite{hasson2020leveraging} & 2020    & 40.72    \\ 
H+O \cite{tekin2019h+}  & 2019   & 40.14      \\ 
H2O \cite{Kwon_2021_ICCV} & 2021    & 39.33      \\ 
HTT \cite{wen2023hierarchical}  & 2023   & 35.33      \\ 
H2OTR \cite{cho2023transformer} & 2023 & 25.10 \\
THOR-Net \cite{aboukhadra2023thor-net} & 2023 & 36.65 \\
SHARP \cite{mucha2025sharp} & 2024 & 28.66    \\ 
\textbf{Ours} & \textbf{2025} & \textbf{22.77}    \\
\bottomrule
\end{tabularx}
\end{table}

\subsection{Dataset selection and cross-dataset evaluation}
\label{supp:sec:cross-domain}

We train our model on \textit{HOT3D}~\cite{banerjee2024hot3d} and evaluate it on the \textit{H2O Dataset}~\cite{Kwon_2021_ICCV} and \textit{AssemblyHands}~\cite{ohkawa:cvpr23}. We select \textit{HOT3D} as the training dataset for two key reasons: (1) it provides the most recent high-quality motion capture data with diverse scene variations and (2) it enables direct comparison with existing methods using \textit{H2O} and \textit{AssemblyHands} for cross-domain evaluation~\cite{wildhands2025,cho2023transformer,hamerpavlakos2024reconstructing,rong2021frankmocap}.
Table~\ref{tab:cross_dataset} presents a comprehensive cross-dataset ablation study demonstrating that V-HPOT consistently improves average performance across all training--testing combinations. Training on \textit{HOT3D} achieves the highest average improvement on unseen domains (\textcolor{darkgreen}{56.5\%}), significantly outperforming \textit{AssemblyHands} (\textcolor{darkgreen}{11.8\%}) or \textit{H2O} (\textcolor{darkgreen}{6.4\%}). This superior performance is a testament to the high quality of the motion capture data and the diverse environmental conditions in \textit{HOT3D}, which provide better domain priors than the smaller \textit{H2O} and \textit{AssemblyHands}, with their strongly distorted monochromatic images.
Our approach focuses specifically on cross-domain generalisation: training on one domain and testing on others. Unlike the state-of-the-art methods presented in Table~\textcolor{wacvblue}{6}, which rely on multiple training domains through purely data-driven approaches, V-HPOT proposes a methodological solution that requires no additional data or labels. This design choice aligns with our goal of improving performance in unseen domains while minimising the quantity of data and domains.

{
\setlength{\tabcolsep}{0pt}
\begin{table}[t]
\small
    \centering
    \caption{\textbf{Cross-dataset 3D pose estimation performance (MPJPE in mm).} \textcolor{darkgreen}{Green}/\textcolor{red}{red} values show percentage improvement/degradation from baseline. The right column displays average improvement across test datasets, demonstrating that V-HPOT consistently improves performance in all cross-dataset scenarios.}
    \label{tab:cross_dataset}
    \begin{tabularx}{\linewidth}{l>{\centering}X>{\centering}p{2.1cm}>{\centering}p{2.1cm}>{\centering\arraybackslash}X}
        \toprule
         & {Train} & \multicolumn{2}{c}{Test} & Avg.~$\Delta$ \\
        \cmidrule(lr){2-2}
        \cmidrule(lr){3-4}
        \cmidrule(lr){5-5}
         & {HOT3D} & {H2O} & {AsHa} & \\
        \midrule
        \textit{base} && 179.6 & 297.7 & -- \\
        \textbf{V-HPOT} && 53.3~{\small\textcolor{darkgreen}{(-70.3\%)}} & 174.5~{\small\textcolor{darkgreen}{(-41.4\%)}} & {\small\textcolor{darkgreen}{-55.9\%}} \\
         \midrule
         & {AsHa} & {H2O} & {HOT3D} & \\
        \midrule
        \textit{base} && 252.1  & 349.6 & -- \\
        \textbf{V-HPOT} && 204.2~{\small\textcolor{darkgreen}{(-19.0\%)}}  & 333.3~{\small\textcolor{darkgreen}{(-4.6\%)}} & {\small\textcolor{darkgreen}{-11.8\%}} \\
        \midrule
         & {H2O} & {AsHa} & {HOT3D} & \\
        \midrule
        \textit{base} &  & 289.1 & 250.8 & -- \\
        \textbf{V-HPOT} &  & 238.8~{\small\textcolor{darkgreen}{(-17.4\%)}} & 262.6~{\small\textcolor{red}{(+4.7\%)}} & {\small\textcolor{darkgreen}{-6.4\%}} \\
        \bottomrule
    \end{tabularx}
\end{table}
}

\subsection{Impact of the testing data during TTO}

In the main paper, we report selecting 5\% of data for our test-time optimisation process during experiments, after which we cease weight updates. Table~\ref{tab:ablation_data_supp} presents performance metrics across varying data quantities. The results demonstrate that our selected 5\% threshold represents an optimal balance for achieving the best absolute 3D pose accuracy across both data domains, and using more of it results in a decrease in performance.

However, selecting a percentage of data is not possible in the real-world online scenario, as the data stream's length is unknown, unlike in our experimental case. Thus, we compare the results when using a fixed data quantity for all datasets, equal to 5\% of the \textit{H2O} data, which is 960 frames. The observation is a marginal decline in performance. 

Our approach, which involves adaptation performed on a relatively small amount of data, only represents a performance trade-off in selecting data quantity and learning rate during test time. We assume that our goal is to achieve the highest performance in the fastest possible time for two reasons: (1) whilst adapting the network faster, more samples result in better predictions, yielding superior final results compared to optimising through the entire test set, and (2) the test-time process has lower computational cost as the optimisation does not process the whole test set for every sample. Due to this, we select a high learning rate ($lr=0.3$) for V-HPOT during TTO compared to training with a scheduled $lr \in <0.1,0.012>$.

The high learning rate makes the TTO process sensitive to outliers that appear more frequently in larger data subsets and affect vulnerable 3D pose understanding, causing instability and excessive optimisation of network weights and biases. Particularly sensitive is the regression component responsible for understanding depth (z-coordinate), explaining the performance decrease. Using a training-range $lr$ eliminates this effect and maintains performance improvements with 100\% data, but increases inference time by $\times5$ for the test set. Our method and learning rate selection, therefore, balance performance gains with computational efficiency.

\subsection{Online vs. Offline TTO}

We compare online versus offline adaptation approaches. In the online method, we continuously update weights until processing 5\% of the data. Conversely, in the offline approach, we load pre-trained weights for each sample and optimise $n$ times specifically for that sample. Table \ref{tab:ablation_offline_online} presents results for the online method alongside offline variants with $n=1, 3$ \text{and} $5$. Our findings indicate that the online approach yields superior performance. Additionally, the offline method incurs substantially higher computational costs as $n$ increases.

\subsection{Root-based vs. absolute regression}

V-HPOT aims to improve absolute depth estimation, which is crucial for specific egocentric applications (e.g., AR/VR and robotics), where metric accuracy is vital for interacting with the physical world. The results demonstrate significant enhancements in absolute 3D pose estimation, while mitigating the root-relative MPJPE-RA error to a lesser extent. Some literature has focused on methods that first estimate hand pose in root-relative space with respect to the wrist joint, e.g.,  InterHand \cite{moon2020interhand2}, and then transfer to absolute values.
To investigate whether the V-HPOT approach can work by first estimating in root-relative space and then transforming to absolute coordinates, we reimplemented our network using a root-relative training strategy before transferring to absolute values.
Table \ref{tab:root_based} compares the original V-HPOT and the RR-based V-HPOT variant. While the RR variant demonstrates greater improvements in MPJPE-RA, all other metrics decline compared to the original V-HPOT, confirming our design choices.

\subsection{Impact of error in initial pose}

Figure \textcolor{wacvblue}{3} shows the pose losses with and without V-HPOT during testing. In the figure, we can observe that despite a wrong initial prediction, V-HPOT improves the final result in most samples. In this study, we dive deeper into the impact of the initial pose.

To evaluate the robustness of our method to inaccurate initial predictions, we conduct a systematic noise augmentation experiment. We add uniform noise sampled from $<\frac{-n}{2},\frac{n}{2}>$
millimetres to the initial 3D hand poses prediction, where $\in \{10, 20, 35, 40, 50\}$. This simulates scenarios where the initial pose estimation contains higher error (MPJPE + noise), testing whether our TTO approach can recover from poor initialisations or if it reinforces incorrect predictions.

Results in Table~\ref{tab:ablation_noise} demonstrate that our approach exhibits degradation rather than catastrophic failure. Higher degradation is observed in \textit{H2O}, as the initial predictions are more accurate than in \textit{AssemblyHands}. The error is higher in the \textit{AssemblyHands} initial prediction, and noise has a lesser impact on it. Even with maximum noise $n=50$, we observe improvements over \textit{base} (not using V-HPOT).

\begin{figure*}[t]
    \centering
    
    \begin{subfigure}[b]{0.805\textwidth}
        \centering
        \includegraphics[width=\textwidth]{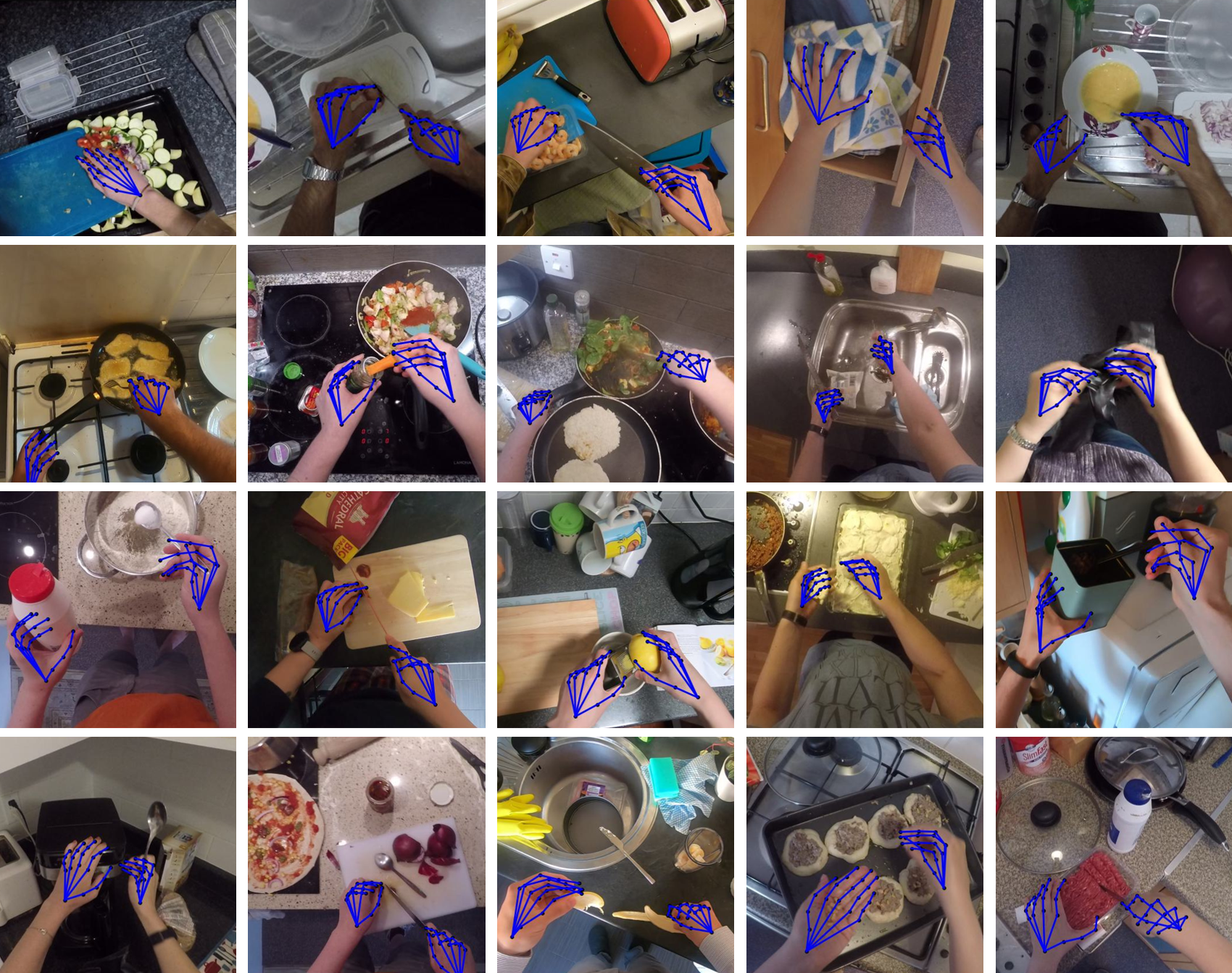}
        \caption{Epick-Kpts}
    \end{subfigure}
    \hfill
    \begin{subfigure}[b]{0.155\textwidth}
        \centering
        \includegraphics[width=\textwidth]{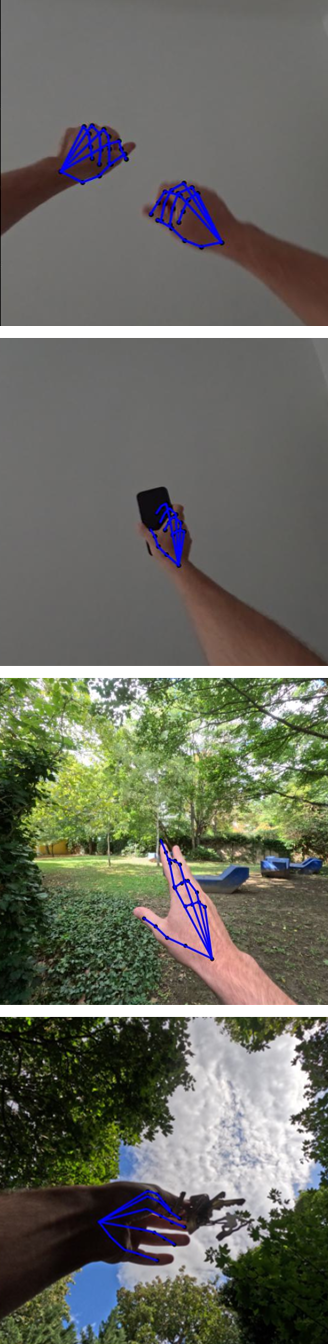}
        \caption{Custom images}
    \end{subfigure}

\caption[Caption for LOF]{\textbf{Qualitative results of our method on the in-the-wild \textit{Epic-Kpts} and own images.} The estimated 3D pose is projected into 2D space, as no ground truth 3D pose labels are available. Despite the challenging natural environments full of everyday objects, various backgrounds, and sceneries, V-HPOT produces geometrically accurate poses. In \textit{Epic-Kpts}, camera wearers perform dynamic actions involving manipulating various objects while preparing meals or cleaning. In addition to \textit{Epic-Kpts}, we capture images without depth cues, featuring an outdoor background and a subject pointing at the wall, without any reference points.
  
  }
  \label{fig:qualitative-epic}
\end{figure*}

{
\setlength{\tabcolsep}{0.5pt}
\begin{table}[t]
\small
    \centering
    \caption{\textbf{Ablation study presenting impact} of test-data quantity considered during our TTO process.}
    \label{tab:ablation_data_supp}
    \begin{tabularx}{\linewidth}{l>{\centering}X>{\centering}X>{\centering}X>{\centering}X>{\centering}X>{\centering}X>{\centering}X>{\centering\arraybackslash}X}
        \toprule
        & \multicolumn{2}{c}{{MPJPE} $\downarrow$} & \multicolumn{2}{c}{{MPJPE-RA} $\downarrow$} & \multicolumn{2}{c}{{MRRPE} $\downarrow$} & \multicolumn{2}{c}{{L2} $\downarrow$} \\
        \cmidrule(lr){2-3}
        \cmidrule(lr){4-5}
        \cmidrule(lr){6-7}
        \cmidrule(lr){8-9}
        & H2O & AsHa & H2O & AsHa & H2O & AsHa & H2O & AsHa \\
        \midrule
        1\% & 125.9 & 217.5 & 67.2 & 98.5 & 65.9 & 344.1 & 5.7 & 51.8 \\
        5\% & 53.3 & \textbf{174.5} & 51.1 & 90.6 & \textbf{54.1} & 253.2 & 5.8 & \textbf{36.9} \\
        7\% & \textbf{52.8} & 184.9 & 51.2 & \textbf{80.42} & 59.3 & \textbf{185.01} & 5.2 & 41.2 \\
        10\% & 74.8 & 184.1 & \textbf{47.1} & 84.6 & 72.2 & 228.5 & \textbf{5.1} & 40.25 \\
        \midrule
        fixed & 53.3 & 175.1 & 51.1 & 92.1 & \textbf{54.1} & 252.9 & 5.8 & 37.0 \\
        \bottomrule
    \end{tabularx}
\end{table}
}
{
\setlength{\tabcolsep}{0.5pt}
\begin{table}[t]
\small
    \centering
    \caption{\textbf{Ablation study presenting impact} of online and offline TTO process strategy selection.}
    \label{tab:ablation_offline_online}
    \begin{tabularx}{\linewidth}{l>{\centering}X>{\centering}X>{\centering}X>{\centering}X>{\centering}X>{\centering}X>{\centering}X>{\centering\arraybackslash}X}
        \toprule
        & \multicolumn{2}{c}{{MPJPE} $\downarrow$} & \multicolumn{2}{c}{{MPJPE-RA} $\downarrow$} & \multicolumn{2}{c}{{MRRPE} $\downarrow$} & \multicolumn{2}{c}{{L2} $\downarrow$} \\
        \cmidrule(lr){2-3}
        \cmidrule(lr){4-5}
        \cmidrule(lr){6-7}
        \cmidrule(lr){8-9}
        & H2O & AsHa & H2O & AsHa & H2O & AsHa & H2O & AsHa \\
        \midrule
        $n=1$ & 153.3 & 306.2 & 66.3 & 105.3 & 74.3 & 277.7 & 6.4 & 96.9 \\
        $n=3$ & 154.4 & 305.7 & 66.7 & 105.3 & 75.0 & 278.4 & 6.3 & 96.6 \\
        $n=5$ & 154.1 & 305.9 & 66.8 & 105.3 & 75.1 & 278.6 & 6.4 & 96.6 \\
        $Online$ & \textbf{53.3} & \textbf{174.5} & \textbf{51.1} & \textbf{90.6} & \textbf{54.1} & \textbf{253.2} & \textbf{5.8} & \textbf{36.9} \\
        \bottomrule
    \end{tabularx}
\end{table}
}

{
\setlength{\tabcolsep}{0.5pt}
\begin{table}[t]
\small
    \centering
    \caption{\textbf{Root-based vs. absolute regression.} Comparison of V-HPOT with regression of absolute coordinates against approach where regressions is done relative to wrist and then transformed to absolute coordinate system.}
    \label{tab:root_based}
    \begin{tabularx}{\linewidth}{l>{\centering}X>{\centering}X>{\centering}X>{\centering}X>{\centering}X>{\centering}X>{\centering}X>{\centering\arraybackslash}X}
        \toprule
        & \multicolumn{2}{c}{{MPJPE} $\downarrow$} & \multicolumn{2}{c}{{MPJPE-RA} $\downarrow$} & \multicolumn{2}{c}{{MRRPE} $\downarrow$} & \multicolumn{2}{c}{{L2} $\downarrow$} \\
        \cmidrule(lr){2-3}
        \cmidrule(lr){4-5}
        \cmidrule(lr){6-7}
        \cmidrule(lr){8-9}
        & H2O & AsHa & H2O & AsHa & H2O & AsHa & H2O & AsHa \\
        \midrule
        \multicolumn{9}{c}{Root-based approach} \\
        \midrule
        base    & 133.5     &  299.7 & 58.1     & 102.5     & 64.2      & 268.7  & 6.6      & 87.1         \\
        V-HPOT  & 57.1      &  209.1 & \textbf{40.5}     & 91.7     & 57.2      & 278.5  & 6.3      & 47.2      \\
        $\Delta$  & \small\textcolor{darkgreen}{57.3\%}    & \small\textcolor{darkgreen}{30.2\%}  & \small\textcolor{darkgreen}{30.2\%}   & \small\textcolor{darkgreen}{10.5\%}     & \small\textcolor{darkgreen}{10.9\%}    & \small\textcolor{red}{3.7\%}  & \small\textcolor{darkgreen}{4.2\%}    & \small\textcolor{darkgreen}{45.8\%}  \\
        \midrule 
        \multicolumn{9}{c}{Absolout depth approach} \\
        \midrule 
        \textit{base} & 179.6 & 297.7 & 52.7 & 105.3 & 126.2 & 301.8 & 7.4 & 80.1 \\
        V-HPOT  & \textbf{53.3} & \textbf{174.5} & 51.1 & \textbf{90.6} & \textbf{54.1} & \textbf{253.2} & \textbf{5.8} & \textbf{36.9} \\
       $\Delta$  & \small\textcolor{darkgreen}{70.3\%}    & \small\textcolor{darkgreen}{41.2\%}  & \small\textcolor{darkgreen}{3.0\%} & \small\textcolor{darkgreen}{12.5\%}     & \small\textcolor{darkgreen}{57.2\%}    & \small\textcolor{darkgreen}{16.2\%}  & \small\textcolor{darkgreen}{21.8\%}    & \small\textcolor{darkgreen}{53.9\%}  \\
        \bottomrule
    \end{tabularx}
\end{table}
}

{
\setlength{\tabcolsep}{0.5pt}
\begin{table}[t]
    \centering
    \small
    \caption{\textbf{Impact of error in initial pose prediction} on V-HPOT performance. Results shown for different noise levels (in mm) applied at TTO.}
    \label{tab:ablation_noise}
    \begin{tabularx}{\linewidth}{l>{\centering}X>{\centering}X>{\centering}X>{\centering}X>{\centering}X>{\centering}X>{\centering}X>{\centering\arraybackslash}X}
        \toprule
        & \multicolumn{2}{c}{{MPJPE} $\downarrow$} & \multicolumn{2}{c}{{MPJPE-RA} $\downarrow$} & \multicolumn{2}{c}{{MRRPE} $\downarrow$} & \multicolumn{2}{c}{{L2} $\downarrow$} \\
        \cmidrule(lr){2-3}
        \cmidrule(lr){4-5}
        \cmidrule(lr){6-7}
        \cmidrule(lr){8-9}
        & H2O & AsHa & H2O & AsHa & H2O & AsHa & H2O & AsHa \\
        \midrule
        \textit{base} & 179.6 & 297.7 & 52.7 &105.3 &126.2 &301.8 &7.4 &80.1 \\
        \textit{V-HPOT} & 53.3 & 174.5 & 51.1 & 90.6 & 54.1 & 253.2 & 5.8 & 36.9 \\
        \midrule
        \textit{n=10mm} & 56.3 & 185.8 & 51.6 & 92.3 & 53.0 & 229.0 & 5.6 & 44.1 \\
        \textit{n=20mm} & 71.5 & 185.4 & 54.8 & 92.3 & 52.0 & 228.3 & 5.6 & 44.0 \\
        \textit{n=30mm} & 94.8 & 190.8 & 58.6 & 90.6 & 53.5 & 225.5 & 5.7 & 46.0 \\
        \textit{n=40mm} & 122.5 & 201.5 & 62.9 & 89.5 & 60.6 & 223.3 & 5.7 & 48.7 \\
        \textit{n=50mm} & 149.2 & 211.2 & 67.2 & 88.4 & 69.1 & 218.0 & 5.7 & 50.3 \\
       
        \bottomrule
    \end{tabularx}
\end{table}
}

\subsection{Qualitative results in Epic-Kpts}

Datasets for egocentric 3D hand pose are limited to a laboratory environment due to annotation methods relying either on a multi-view camera setup or motion capture technology. This limits the recording environments, as all data is confined to indoor laboratory scenes. In contrast, hand pose estimation aims to work in an in-the-wild setup with a more diverse environment. We show qualitative results of our method on the real-world dataset \textit{Epic-Kpts}~\cite{wildhands2025}, which is a subset of the \textit{Epic-Kitchens}~\cite{Damen2018EPICKITCHENS} dataset. Captured in the various kitchens where subjects prepare meals, it is full of natural actions, motion blur, and a wide variety of interacting objects. Unfortunately, there are no 3D pose labels for comparison, limiting us to qualitative analysis only. In addition to Epic-Kpts, we capture images with different background structures where hands are in front of a wall without any reference information, and outdoors in the garden, where the background is far from the camera. Results in Figure~\ref{fig:qualitative-epic} show that our V-HPOT result in accurate pose estimates even in such challenging scenes.

\subsection{Qualitative analysis of pseudo-depth estimation models}
\label{supp:sec:qual}

\begin{figure}[t]
    \centering
    
    \begin{subfigure}[b]{0.235\textwidth}
        \centering
        \includegraphics[width=\textwidth]{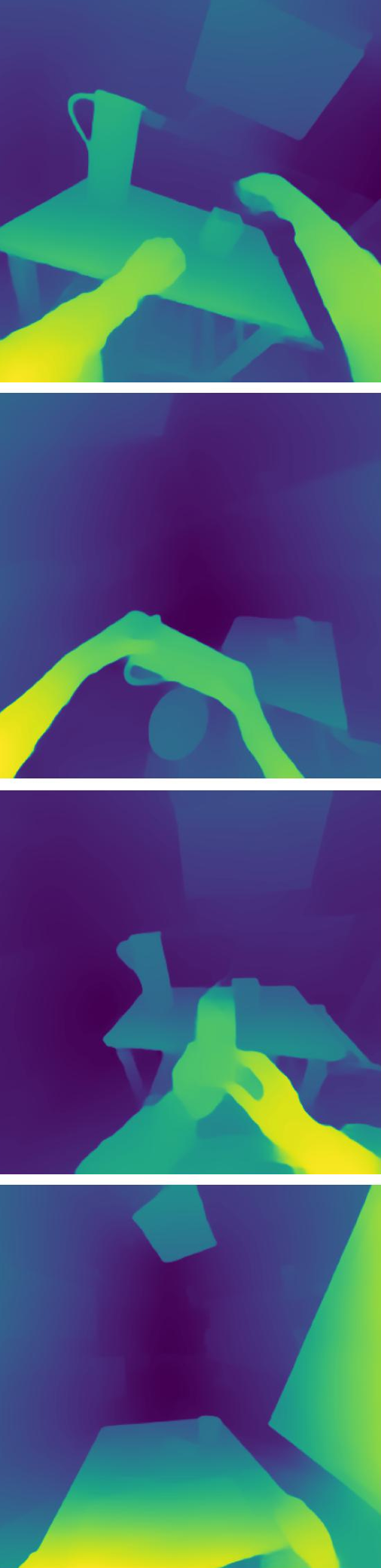}
        \caption{DPT-Hybrid}
    \end{subfigure}
    \hfill
    \begin{subfigure}[b]{0.235\textwidth}
        \centering
        \includegraphics[width=\textwidth]{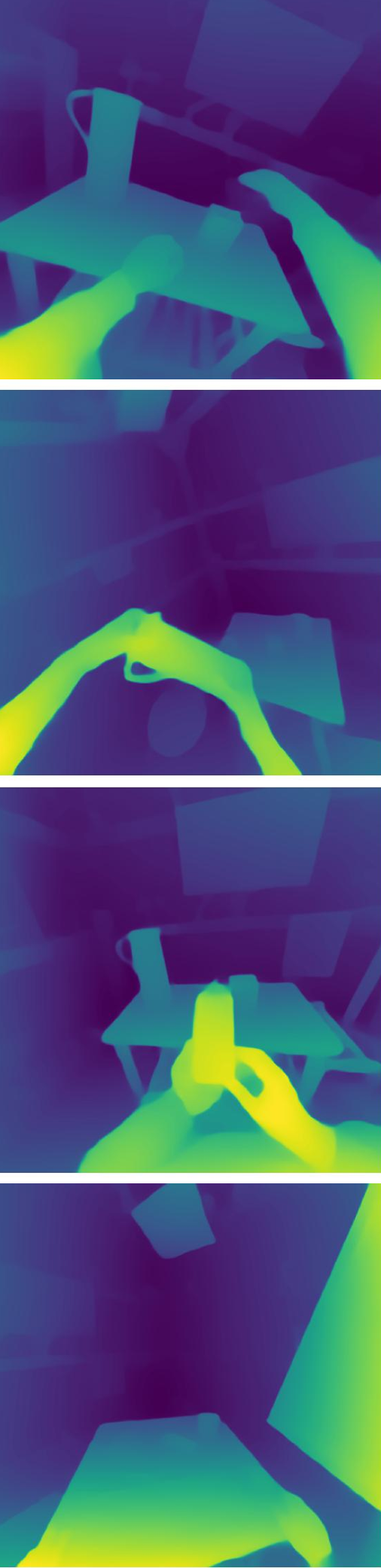}
        \caption{DepthAnything}
    \end{subfigure}
    \caption{A visual comparison of the \textit{DPT-Hybrid} and \textit{DepthAnything} pseudo-depth estimation models. \textit{DepthAnything} is better at estimating background details, while the quality of hand estimates is similar in both methods.}
    \label{fig:depth-anything-vs-midas}
\end{figure}

In the main paper, we compare two pseudo-depth estimation models: \textit{DPT-Hybrid}~\cite{ranftl2021vision} and \textit{DepthAnything}~\cite{yang2024depth}. The results vary depending on the metric used. We measured inference using an NVIDIA RTX 3090 GPU over 1000 trials. Given the mixed quantitative results, \textit{DPT-Hybrid} is faster (26.8 ms vs. 39.8 ms), which is essential at the experimental stage. Furthermore, we visually analyse the differences between \textit{DPT-Hybrid} and \textit{DepthAnything} by imagining random frames from the \textit{HOT3D} dataset. These frames are presented in Figure \ref{fig:depth-anything-vs-midas}. We observe that both models perform similarly in terms of hand regions. The main difference is \textit{DepthAnything}'s superior understanding of the background details. In our study, which focuses solely on the hands, these background details are less important, while the faster inference and better absolute 3D error in pose estimation favour \textit{DPT-Hybrid} for the role of our auxiliary task.

\subsection{Ground truth depth vs. pseudo-depth analysis}

Pseudo-depth estimation, used as an auxiliary task, is limited by errors that can propagate during hand pose estimation training. Among the datasets in this study, only \textit{H2O} provides sensor-based ground-truth depth measurements, limiting the possible cross-domain experiments. To evaluate the impact of depth estimation quality, we train our network on the \textit{H2O} dataset and assess its performance in a cross-domain scenario on \textit{HOT3D} and \textit{AssemblyHands}, using ground-truth depth and \textit{DPT-Hybrid} pseudo-depth as auxiliary tasks.

Results reveal a negligible difference for \textit{base} (without TTO) (0.3\% MPJPE) between GT depth versus pseudo-depth when averaged across both domains. With TTO applied, the results are mixed. For \textit{HOT3D}, the difference in  MPJPE is minimal in favour of the pseudo-depth (262.6 vs. 263.2), while for AssemblyHands is larger (238.8 vs 250.7). This suggests that pseudo-depth estimation provides sufficient spatial correctness to serve as an auxiliary task. In some cases, it even improves performance through a regularisation effect. The less detailed background information in slightly imperfect pseudo-depth may prevent overfitting to specific depth patterns coming from the scene background or sensor noise, but being not crucial for hand understanding, leading to better generalisation across domains and serving effectively as an auxiliary task. Additionally, pseudo-depth remains advantageous as an auxiliary task because it can be used on any dataset without requiring a depth sensor. 

\end{document}